\documentclass[final,5p,times,twocolumn]{elsarticle}
\usepackage{amsmath,amssymb}
\usepackage{graphicx}
\usepackage{booktabs}
\usepackage{xcolor}
\usepackage{listings}
\usepackage{tikz}
\usetikzlibrary{arrows.meta,positioning}
\usepackage{url}
\journal{Energy and AI}
\biboptions{sort&compress}
\begin{document}
\begin{frontmatter}
\title{Interpretable Cyber-Physical Attack Detection in Smart Grids: A Leakage-Controlled Multi-Testbed Evaluation of Fusion, Robustness, and Observability Limits}
\author[inq,ece]{Muhammad Ridwanul Hoque}
\ead{ridwanalvee@gmail.com}
\author[inq,ece]{Sarah Sharif}
\author[ece]{Reza Saeed Kandezy}
\author[inq,ece]{Yaser Banad\corref{cor1}}
\ead{bana@ou.edu}
\cortext[cor1]{Corresponding author.}
\affiliation[inq]{organization={Intelligent Neuromorphic and Quantum Understanding for Innovative Research and Engineering (INQUIRE) Laboratory},city={Norman},postcode={73019},state={Oklahoma},country={U.S.A.}}
\affiliation[ece]{organization={School of Electrical and Computer Engineering, University of Oklahoma},city={Norman},postcode={73019},state={Oklahoma},country={U.S.A.}}
\begin{abstract}
Cyber-physical smart-grid detectors are typically treated as black boxes and validated exclusively on clean, temporally shuffled datasets, which risk overly optimistic performance claims. This work presents a multi-testbed evaluation study built around an interpretable, time-windowed fusion Random Forest, adopting the pessimistic result whenever evaluation protocols disagree. (i) On a public paired cyber-physical dataset, shuffled cross-validation is optimistic: macro-F1 falls from $0.96$ to $0.88\pm0.06$ under rolling-origin temporal evaluation. A topology-free graph convolutional network (GCN) serving as a control confirms that this lightweight model is competitive on these splits. (ii) The adaptive and adaptive-joint feature-space attacks, which are infeasible as realistic grid manipulations, degrade macro-F1 to $0.67$ and $0.34$ respectively. Adversarial training and randomized smoothing partially restore performance, and a physics-constrained evolution-strategy evasion on the IEEE 14-Bus testbed shows the same vulnerability pattern. (iii) On real IEC-104 traffic, the cyber view reliably catches structural attacks (recall of $0.87$--$0.98$) but largely fails on semantic value manipulation (recall of $0.11$--$0.35$). An observability taxonomy on the IEEE 14-Bus system reveals that stealthy false data injection defeats all three observation views. (iv) On a full AC weighted least-squares state-estimation testbed, a forecasting-aided temporal test is evaded by slow ramps precisely where a rate-invariant physics-manifold test sustains detection at a common $5\%$ false-alarm target. A full end-to-end evaluation on IEEE 118-Bus confirms the ordering but bounds its strength: the manifold test detects every trial down to $0.05^\circ$/step and $0.48$ at $0.02^\circ$/step, against $0.14$ for the temporal test. All code and benchmarks are released.
\end{abstract}
\begin{keyword}
smart grid security\sep false data injection\sep adversarial machine learning\sep explainable AI\sep state estimation\sep cyber-physical fusion
\end{keyword}
\end{frontmatter}

\section{Introduction}
The electric power grid has evolved into a cyber-physical system. Wide-area monitoring, protection, and control functions, once served by legacy transducer-based instrumentation, depend on phasor measurement units (PMUs), programmable logic controllers (PLCs), remote terminal units (RTUs), and Supervisory Control and Data Acquisition (SCADA) software communicating over standardized protocols such as IEC~60870-5-104~\cite{iec60870} and IEC~61850~\cite{iec61850}. This digitization has delivered faster control, better observability, and the flexibility needed to integrate renewable and distributed resources. It has also fused two historically separate attack surfaces, the information-technology (IT) network and the physical apparatus, into one. The 2015 and 2016 attacks on the Ukrainian power grid~\cite{case2016ukraine} showed concretely that a purely digital intrusion can cascade into large-scale physical outages, and subsequent analyses have framed grid security as an inherently cyber-physical problem~\cite{musleh2020survey,pasqualetti2013attack}.

Threats accordingly span both layers. On the cyber side, adversaries reuse mature IT techniques such as credential brute-forcing, backdoors, reverse shells, and ransomware to gain and retain control of grid devices. On the physical side, the archetypal threat is the false data injection (FDI) attack, in which an adversary with knowledge of the system topology crafts measurement manipulations that satisfy the power-flow constraints and therefore evade classical residual-based bad-data detection~\cite{liu2009fdi,liu2011fdi}. Full and current topology knowledge is the strongest adversarial assumption and the standard one in the FDI literature; because acquiring it in practice is difficult, the graded-knowledge sweep of Section~\ref{sec:resD3} relaxes it to the partial-knowledge case. The two layers are tightly coupled: a software compromise is frequently the delivery mechanism for a physical manipulation, while the physical consequence (an incorrect state estimate or a misoperated breaker) is what ultimately threatens stability. Detecting the complete attack therefore benefits from observing the network telemetry and the physical measurements jointly, a paradigm known as cyber-physical fusion~\cite{atse2015fusion,sweeten2025fusion}. Recent research has produced increasingly powerful detectors on both the attack and defense sides, from generative deep-learning FDI in renewable-integrated grids~\cite{dlfdi2025} to graph- and transformer-based intrusion detectors~\cite{graphkan2025,gnntrans2025}.

Despite this progress, three gaps persist. First, the strongest fusion detectors are opaque deep or graph neural networks that offer little insight into why an alarm was raised, hindering operator trust and root-cause analysis. Second, and more fundamentally, detectors are almost always evaluated only on clean data; devices are certified against corrupted data and electromagnetic interference under existing standards, but not against deliberate adversarial manipulation, a striking omission in a security setting whose premise is an intelligent, adaptive adversary. The broader machine-learning community has repeatedly shown that classifiers can be fooled by small, bounded perturbations that transfer between models~\cite{szegedy2014intriguing,goodfellow2015fgsm,papernot2016transferability}, and that robustness demands deliberate defenses~\cite{madry2018pgd,cohen2019certified,mtd2025}. Third, single-testbed evaluations leave open whether results generalize; yet paired cyber-physical attack datasets are rare, and where they exist they may lack a grid topology, precluding physically-faithful attack construction.

This paper addresses all three gaps with a leakage-controlled, multi-testbed study organized around a single interpretable detector. A lightweight Random Forest over fused, time-windowed features is evaluated under four protocols of increasing stringency and against four other model families, and it remains fully explainable through SHAP (SHapley Additive exPlanations)~\cite{lundberg2017shap}. A topology-free graph control is reported in \ref{app:gcn}; the topology-aware fusion GNN of~\cite{sweeten2025fusion} is not reimplemented. The detector is then subjected to a graded battery of adversarial attacks in measurement space (random, gradient-transfer, adaptive, and joint) and defended with adversarial training and randomized smoothing. Because the primary dataset exposes no grid topology, a second testbed is built, an IEEE 14-Bus system in \texttt{pandapower}~\cite{thurner2018pandapower} (whose case data follow the standard MATPOWER benchmarks~\cite{zimmerman2011matpower}), on which physics-feasible stealthy FDI is constructed and attacked with a population-based evolution strategy. The cyber side is cross-validated on independent real IEC-104 traffic, where five behavioral attacks and a command-transition anomaly model establish a principled boundary of cyber-only detection. Finally, the physics testbed is extended to a full AC state estimator with temporal and spatial consistency detectors, and the comparison is repeated end to end on IEEE 118-Bus.

The contributions support two claims. The first two support Claim~1, that shuffled, clean evaluation overstates security performance. The last two support Claim~2, that detectability is set by the observable rather than by the model.
\begin{itemize}
\item a leakage-controlled evaluation of an interpretable cyber-physical fusion detector under four protocols (shuffled, time-blocked, rolling-origin, and chronological hold-out) and five model families, showing that shuffling inflates every family by more than the non-linear families differ from one another under honest evaluation;
\item a graded adversarial robustness study that separates feature-space worst cases (with adversarial-training and smoothing defenses) from physically feasible attacks, the latter constructed on an IEEE 14-Bus \texttt{pandapower} testbed with a strong evolution-strategy evasion;
\item a cyber-only detection boundary established on real IEC-104 traffic and formalized as a semantic-manipulation observability taxonomy on IEEE 14-Bus that localizes which observation layer each attack class requires; and
\item a study of temporal versus spatial detectors on a full AC weighted least-squares (WLS) state-estimation testbed, validated end to end on IEEE 118-Bus, showing that a Jacobian-consistent stealthy FDI is invisible to snapshot bad-data detection, that a rate-invariant physics-manifold test closes the slow-ramp evasion that defeats forecasting-aided detection, and where that test stops being reliable.
\end{itemize}

\emph{Scope of claims.} Fusion is evaluated across three complementary testbeds rather than on a single end-to-end co-simulated system: dataset~1 provides paired real cyber-physical captures without grid topology; the IEEE 14-Bus and IEEE 118-Bus testbeds provide physics with a simulated cyber layer; and dataset~3 provides real traffic without physics. Every claim below is scoped to the testbed that supports it, and the residual gap, an aligned real-traffic real-topology co-simulation, is stated as future work (Section~\ref{sec:disc}). This scope statement is made once here and is not repeated. The paper is organized around two claims. Section~\ref{sec:rel} reviews related work. Section~\ref{sec:method} sets out the theory: what each observation layer carries, state estimation and stealthy FDI, the consistency detectors, the evaluation protocol, the threat models and defenses, and the classifier. Section~\ref{sec:setup} describes the test scenarios. Section~\ref{sec:results} reports the results in four groups, each supporting one claim, and Sections~\ref{sec:disc} and~\ref{sec:conc} discuss the results and state the claims.

\section{Related Work}\label{sec:rel}
Five threads bear on this study; each closes with the gap this paper addresses.
\subsection{FDI and its detection}
Stealthy FDI was formalized by Liu, Ning and Reiter~\cite{liu2009fdi,liu2011fdi}; the classical theory and practice of weighted least-squares state estimation and bad-data detection follow Monticelli~\cite{monticelli2000se} and Abur and G\'omez Exp\'osito~\cite{abur2004power}; ML detectors span SVM/sparse methods~\cite{ozay2016ml,esmalifalak2017detecting}, deep learning~\cite{he2017realtime,dlfdi2025}, transfer and self-supervised learning~\cite{deeptransfer2021,selfsup2024}, and graph/transformer networks~\cite{boyaci2022gnn,graphkan2025,gnntrans2025}; comprehensive surveys~\cite{musleh2020survey,sayghe2020survey,fdisurvey2023}; public benchmarks~\cite{hink2014discrimination,matousek2022ics}; and the theoretical limits of CPS attack detection~\cite{pasqualetti2013attack}.
\subsection{Cyber-physical fusion}
Traffic-indexed state estimation~\cite{atse2015fusion}; the topology-aware GNN fusion detector of Sweeten et al.~\cite{sweeten2025fusion} on the dataset we use. We do not reproduce their architecture; we contribute an interpretable baseline under leakage-controlled protocols, plus robustness, reliability, and cross-validation, and our Section~\ref{sec:resD4} takes up their stated future work on slow injection attacks.
\subsection{Adversarial machine learning}
The fast gradient sign method (FGSM)~\cite{goodfellow2015fgsm} and projected gradient descent (PGD)~\cite{madry2018pgd}; certified smoothing~\cite{cohen2019certified}; moving-target defenses~\cite{mtd2025}; distillation shown insufficient~\cite{papernot2016distillation,carlini2017robustness}; transferability~\cite{papernot2016transferability}; grid adversarial FDI~\cite{lesson2024,advresilient2021,phantom2025}.
\subsection{Temporal and subspace FDI detection}
Beyond snapshot residuals, forecasting-aided state estimation~\cite{zhao2017fase} and Kalman/CUSUM innovation tests~\cite{manandhar2014kalman} exploit temporal dynamics, while low-rank/subspace methods exploit the spatial structure of benign states~\cite{liu2014sparse}; each has known evasions, notably manifold-aligned injections that bypass low-rank detection~\cite{zhang2017bypass}. We relate these two threads on one AC testbed (Section~\ref{sec:resD4}).
\subsection{Explainability and selective prediction}
LIME~\cite{ribeiro2016lime}; SHAP~\cite{lundberg2017shap}; recent grid XAI~\cite{xai2026}. Selective/abstaining prediction augments a classifier with a reject option; we use it as an escalation policy to the physical layer.

\section{Theory and Methodology}\label{sec:method}
\emph{Notation.} Bold upper-case symbols denote matrices and bold lower-case symbols vectors; $\log$ is the natural logarithm. The bold $\mathbf{H}$ is always the measurement Jacobian, while the calligraphic $\mathcal{H}$ is always a Shannon entropy, of command symbols in Eq.~\ref{eq:cmd} and of the predictive distribution in Section~\ref{sec:clf}; $\mathcal{L}$ is a classification loss. The symbol $\sigma$ always denotes a standard deviation, and its subscript states of which quantity. The code listings keep the variable names of the released scripts.

\subsection{Observables and features}\label{sec:data}
Each observation layer carries a different physical quantity, and this sets what any detector built on it can see. Network headers carry the structure and timing of the traffic: which devices communicate, how often, and with which protocol and message type. Payloads carry the values being reported, such as a measured power flow or a commanded setpoint. Physical measurements carry the state of the grid, and through the power-flow model they must be consistent with one another. A detector can only recover what is present in the layer it reads. A value manipulated inside the payload is therefore absent from header features, and no function of those features can recover it (Section~\ref{sec:resD3}). This framing is used throughout the paper: each attack class is matched to the layer in which its effect is actually present.

The study uses three data sources, referred to throughout as \emph{dataset~1} (paired real cyber-physical captures), \emph{testbed~2} (the simulated IEEE 14-Bus system; its real-time AC extension is testbed~2$'$), and \emph{dataset~3} (real IEC-104 traffic); ``dataset'' denotes measured data, ``testbed'' a simulation.

\emph{Dataset~1:} \mbox{SmartGrid-CyberPhysical-Attack-Dataset}~\cite{sweeten2025fusion,sweeten2025dataset}, six classes, paired \texttt{Cyber.csv}/\texttt{Physical.csv}. The raw captures contain millions of packet and measurement rows; our unit of analysis is the $\Delta=0.5$\,s \emph{window}, and a window survives only if the cyber and physical streams overlap in that bin (inner join on the time bin). Because each class capture spans only minutes of wall-clock time, this windowing-and-join yields $1{,}529$ fused windows in total ($204$--$352$ per class: backdoor $204$, brute-force $216$, FDI $251$, ransomware $252$, reverse-shell $254$, benign $352$); per-window sample sizes for every split are reported with the results. Each class is recorded as a separate capture, which is treated as a threat to validity in Section~\ref{sec:disc}. This is far fewer samples than the packet-level counts of~\cite{sweeten2025fusion} operate on, which is why we treat interval estimates (Section~\ref{sec:resD1}) rather than single splits as the conservative temporal number. \emph{Testbed~2:} IEEE 14-Bus in \texttt{pandapower}~\cite{thurner2018pandapower}; feasible $a=Hc$ injections. \emph{Dataset~3:} real IEC 60870-5-104~\cite{iec60870} and IEC 61850/MMS traces~\cite{matousek2022ics} from real and virtual substations, with labeled attacks. Table~\ref{tab:feats} lists the dataset-1 features.

\begin{table}[t]\centering\caption{Engineered per-window features, dataset 1 ($\Delta=0.5$\,s).}\label{tab:feats}
\footnotesize\setlength{\tabcolsep}{3pt}\begin{tabular}{ll}\toprule Domain & Features\\\midrule
Physical (22) & mean/std of Freq, Theta, $V_{A,B,C}$, $I_{A,B,C}$,\\
 & active/reactive power, breaker status\\
Cyber (12) & packet count; mean/sum/std IP length; \#unique\\
 & src/dst IPs; fraction of TCP, Modbus/TCP, HTTP,\\
 & TLS, ARP, ICMPv6\\\bottomrule\end{tabular}\end{table}

Both streams are binned into windows of width $\Delta$. For window $w$ and physical measurement set $M$,
\begin{equation}\mathbf{x}_w^{P}=\big[\mu_m,\sigma_m\big]_{m\in M},\quad \mu_m=\tfrac{1}{|w|}\!\sum_{t\in w}m_t,\quad \sigma_m=\operatorname{std}_{t\in w}(m_t),\end{equation}
where $|w|$ is the number of raw samples falling in window $w$ and $m_t$ the value of measurement $m$ at time $t$.
Cyber features aggregate the packets ($b=$ IP length in bytes, $\pi_p=$ fraction of packets using protocol $p$):
\begin{equation}\mathbf{x}_w^{C}=\big[n_{\text{pkt}},\mu(b),\textstyle\sum b,\sigma_b,|{\rm src}|,|{\rm dst}|,\{\pi_p\}_{p\in P}\big].\end{equation}
The fused representation concatenates the domains (\emph{fusion} is used in the feature-level, early-fusion sense throughout: the classifier learns cross-domain interactions through joint tree splits, but no latent joint representation is learned; see Section~\ref{sec:disc}):
\begin{equation}\mathbf{x}_w=\big[\mathbf{x}_w^{P}\oplus\mathbf{x}_w^{C}\big]\in\mathbb{R}^{d},\end{equation}
with $d=34$ ($22$ physical $+$ $12$ cyber features); the feature dimension $d$ is distinct from the state dimension $n$ of Eq.~\ref{eq:se}.

For IEC-104 we model command \emph{semantics}. Let $s_t=(\text{asduType}_t,\text{CoT}_t)$ be the command symbol with empirical benign distribution $\hat P(s)$ over support $\mathcal{S}$. Per window $w$ a surprise $\zeta$, a novelty $\nu$, and a symbol entropy $\mathcal{H}$ are computed ($\zeta$ rather than $\Sigma$, which is reserved for summation):
\begin{align}\label{eq:cmd}
\zeta(w)&=\tfrac{1}{|w|}\!\sum_{t\in w}\!-\log\hat P(s_t),\qquad \nu(w)=\tfrac{1}{|w|}\!\sum_{t\in w}\!\mathbf{1}[s_t\notin\mathcal{S}],\nonumber\\
\mathcal{H}(w)&=-\!\sum_{s}\hat p_w(s)\log\hat p_w(s).
\end{align}

\subsection{State estimation and stealthy FDI}\label{sec:se}
State estimation and its bad-data detection are standard practice~\cite{monticelli2000se,abur2004power}. Under the standard (DC) measurement model, meter readings $\mathbf{z}\in\mathbb{R}^m$ relate to the state $\mathbf{x}\in\mathbb{R}^n$ (bus angles) through the measurement Jacobian $\mathbf{H}\in\mathbb{R}^{m\times n}$ and zero-mean measurement noise $\mathbf{e}$:
\begin{equation}\label{eq:se}\mathbf{z}=\mathbf{H}\mathbf{x}+\mathbf{e},\qquad \hat{\mathbf{x}}=(\mathbf{H}^\top\mathbf{W}\mathbf{H})^{-1}\mathbf{H}^\top\mathbf{W}\mathbf{z},\end{equation}
where $\mathbf{W}$ is the diagonal matrix of inverse measurement variances and $\hat{\mathbf{x}}$ the weighted least-squares state estimate; bad data are flagged when the residual exceeds a threshold set by the assumed noise level, $\lVert\mathbf{z}-\mathbf{H}\hat{\mathbf{x}}\rVert_2>\eta$. A stealthy FDI attack~\cite{liu2011fdi} adds $\mathbf{a}=\mathbf{H}\mathbf{c}$, where $\mathbf{c}\in\mathbb{R}^n$ is an attacker-chosen state offset, giving $\mathbf{z}_a=\mathbf{H}(\mathbf{x}+\mathbf{c})$ whose residual is unchanged, feasible and undetectable by \eqref{eq:se}.

The AC measurement model maps voltages $V$ and angles $\theta$ to active and reactive power injections via the bus admittance matrix $\mathbf{Y}=\mathbf{G}+j\mathbf{B}$:
\begin{align}
P_i&=V_i\!\sum_k V_k(G_{ik}\cos\theta_{ik}+B_{ik}\sin\theta_{ik}),\label{eq:acp}\\
Q_i&=V_i\!\sum_k V_k(G_{ik}\sin\theta_{ik}-B_{ik}\cos\theta_{ik}).\label{eq:acq}
\end{align}
Under the DC approximation, a linearization valid near the operating point, the measurements are linear in the state, $\mathbf{z}=\mathbf{H}\mathbf{x}+\mathbf{e}$, with $\mathbf{x}$ the bus-angle state of Eq.~\ref{eq:se}, $\mathbf{H}$ built from the branch reactances of the IEEE 14-Bus case, and $\mathbf{e}$ the metering noise. On this testbed the benign snapshots are generated directly from this model: bus angles are drawn at random and mapped through $\mathbf{H}$ with added metering noise. A stealthy attack injects $\mathbf{a}=\mathbf{H}\mathbf{c}$. Because $\mathbf{a}$ lies in the column space of $\mathbf{H}$, the least-squares residual $\mathbf{r}=\mathbf{z}_a-\mathbf{H}\hat{\mathbf{x}}$ is unchanged by the attack. It stays at the metering-noise floor, $3.9{\times}10^{-3}$ for both benign and attacked snapshots (Table~\ref{tab:tax}), and the attack changes it by only about $10^{-14}$, which is numerical zero in double-precision arithmetic. Two clarifications follow. First, this is the residual of the \emph{DC} (linear) estimator against which $\mathbf{a}=\mathbf{H}\mathbf{c}$ is constructed. It is \emph{not} the AC residual that an operator's nonlinear WLS estimator would compute. Second, a DC-designed attack is in general \emph{not} AC-stealthy, because the AC measurement map is nonlinear; Section~\ref{sec:resD4} measures how much topology knowledge a DC-designed attacker needs before the AC chi-square test stops flagging it. Sections~\ref{sec:resD2} and~\ref{sec:resD3} are therefore DC-model constructions, in which both the measurements and the attacks follow the DC model. Section~\ref{sec:resD4} moves to the full AC estimator, with measurements generated by AC power flow and an AC-exact attacker, and exposes the DC--AC gap explicitly.

The static DC construction above is limited to a single snapshot. This subsection extends the testbed to a full nonlinear AC weighted-least-squares (WLS) estimator on the same IEEE 14-Bus grid, driven by time-varying operating points. Each bus load follows a common slow cycle with a bus-specific phase offset, together with an independent first-order autoregressive fluctuation. The operating point at every step is obtained by solving an AC power flow. The estimator minimizes $J(\mathbf{x})=\mathbf{r}^\top\mathbf{W}\mathbf{r}$, $\mathbf{r}=\mathbf{z}-\mathbf{h}(\mathbf{x})$, by Gauss--Newton,
\begin{equation}\label{eq:wls}\mathbf{x}^{k+1}=\mathbf{x}^{k}+(\mathbf{H}^\top\mathbf{W}\mathbf{H})^{-1}\mathbf{H}^\top\mathbf{W}\,\mathbf{r}(\mathbf{x}^k),\ \ \mathbf{H}=\tfrac{\partial\mathbf{h}}{\partial\mathbf{x}},\end{equation}
with $\mathbf{h}$ the AC measurement map (Eqs.~\ref{eq:acp}--\ref{eq:acq}), $m{=}72$ measurements ($|V|$, $P/Q$ injections, $P/Q$ flows) and $n{=}27$ states, giving a measurement redundancy of $m/n\approx2.7$ ($\approx3.0$ on IEEE 118-Bus). This is within the range typical of academic benchmark systems and of the redundancy reported for utility estimators. The measurement mix is representative of conventional SCADA-based AC state estimation. It contains no phasor measurement unit (PMU) data, so the results here describe the SCADA-only case, and extension to hybrid SCADA/PMU estimation~\cite{gol2015hybrid} is left to future work. Measurements are weighted by inverse variance using a single metering-noise level for all measurement types. This is a simplification adopted for this study. A more realistic model assigns distinct variances to voltage, injection and flow measurements, and this is noted as a limitation. Bad data are flagged by the chi-square test $J(\hat{\mathbf{x}})>\chi^2_{m-n,0.99}$ and the largest-normalized-residual test $\max_i|r_i|/\sqrt{\Omega_{ii}}>3$, $\boldsymbol\Omega=\mathbf{R}-\mathbf{H}(\mathbf{H}^\top\mathbf{W}\mathbf{H})^{-1}\mathbf{H}^\top$, where $\mathbf{R}=\mathbf{W}^{-1}$ is the measurement-error covariance. An AC-exact stealthy attacker picks a consistent alternate state $\mathbf{x}_a$ and injects $\mathbf{a}=\mathbf{h}(\mathbf{x}_a)-\mathbf{h}(\mathbf{x})$. This leaves $J$ and every $r_i$ unchanged (Listing~\ref{lst:fdi}), so no residual-based bad-data test can flag it from a single snapshot. A test that compares the estimated state with the benign operating manifold is not residual-based, and it is introduced next.

\begin{lstlisting}[caption={AC-exact stealthy FDI: a consistent alternate state leaves the WLS residual unchanged.},label={lst:fdi}]
def ac_exact_attack(x, bias):    # bias one state coord
    x_a = x.copy(); x_a[tgt] += bias
    return h(x_a) - h(x)         # a = h(x_a) - h(x)
xh_a, H = wls(z_clean + a, x)    # estimator moves to x_a
Ja, _  = bdd(z_clean + a, xh_a, H)   # Ja == Jc (passes BDD)
\end{lstlisting}

\subsection{Consistency detection}\label{sec:consist}
Two consistency detectors are therefore defined on the estimated-state stream $\hat{\mathbf{x}}_t$. They read different axes of the data, and each has a known evasion.

The first is a \emph{temporal} test. It forms a Holt linear-trend one-step prediction $\tilde x_t$ of a monitored coordinate and runs a cumulative sum (CUSUM) on the normalized innovation $\iota_t=(\hat x_t-\tilde x_t)/\hat\sigma_\iota$. This test reads change over time. Its known evasion is a slow ramp. If the bias is introduced slowly enough, the trend filter adapts to it, the innovation returns to the noise floor, and the attack passes unflagged.

The second is a \emph{spatial}, or manifold, test. Legitimate operating points occupy a low-dimensional manifold set by the load variation. The test learns that manifold by principal component analysis (PCA) on benign estimated states and monitors the off-manifold residual
\begin{equation}\label{eq:rho}\rho(\hat{\mathbf{x}})=\big\lVert(\mathbf{I}-\mathbf{P}_K)\,\tilde{\mathbf{q}}\big\rVert_2,\qquad \tilde{\mathbf{q}}=(\hat{\mathbf{x}}-\boldsymbol\mu_x)\oslash\boldsymbol\sigma_x-\bar{\mathbf{q}},\end{equation}
with $\boldsymbol\mu_x$ and $\boldsymbol\sigma_x$ the per-coordinate mean and standard deviation of the benign estimated states, $\mathbf{P}_K$ the projector onto the $K$-dimensional benign subspace and $\bar{\mathbf{q}}$ the mean standardized benign state, subtracted to center the stream (Listing~\ref{lst:rho}). A CUSUM on the standardized $\rho$ raises the alarm. This test reads distance from the benign subspace rather than change over time. Because $\rho$ depends on the injected \emph{bias} and not on how slowly that bias is introduced, it is rate-invariant. Its own known evasion is an injection that stays on the manifold, which a coordinated multi-bus attacker can approach.

Two choices in fitting the manifold matter, and both are made conservatively. First, the manifold is fitted on the estimated states that an operator observes, not on the true power-flow states. Fitting on true states would supply ground-truth information that is not available in deployment, and it produces an optimistically clean subspace. Second, the dimension $K$ is selected by cross-validation on benign data alone, with no access to attacked data. A cumulative-variance rule is not used. Applied to estimated states, such a rule returns $K{=}20$ of $27$ on this testbed and $K{=}165$ of $235$ on IEEE 118-Bus, where it gives a badly miscalibrated detector. Cross-validation returns $K{=}6$ here and $K{=}8$ on IEEE 118-Bus.

The low dimension is a property of the system rather than of the load model. When the load profiles are regenerated with an independent frequency for every bus, the benign spectrum is almost unchanged, and the dimension needed to reach $99.9\%$ of the variance moves only from $5$ to $6$. This indicates that the AC power-flow mapping itself concentrates independent load variation into a low-dimensional state response.

Both detector thresholds are calibrated on independent benign streams to a common $5\%$ false-alarm target, and the rate each test actually achieves on held-out benign streams is reported with every result. The \emph{fused} detector takes the normalized maximum of the two.

\begin{lstlisting}[caption={Physics-manifold consistency: the off-manifold residual is rate-invariant.},label={lst:rho}]
mu, sd = Xb.mean(0), np.maximum(Xb.std(0), 5e-3)
_,_,Vt = np.linalg.svd((Xb-mu)/sd - Zmean, full_matrices=False)
Vk = Vt[:K]                      # K-dim benign load manifold
def rho(x):
    z = (x-mu)/sd - Zmean
    return np.linalg.norm(z - z @ Vk.T @ Vk)  # off-manifold
\end{lstlisting}

We are careful about the novelty claimed here. Forecasting-aided state estimation~\cite{zhao2017fase} and Kalman/CUSUM innovation tests~\cite{manandhar2014kalman} are established, as is subspace/low-rank FDI detection~\cite{liu2014sparse}; an attacker who coordinates a multi-bus injection to stay on the benign manifold can bypass a low-rank test~\cite{zhang2017bypass}, precisely the residual limit of $\rho$. Our contribution is not a new primitive but a \emph{unification}: temporal and spatial consistency fail on orthogonal attack regimes (slow-ramp vs.\ manifold-aligned), and on one AC testbed at a common false-alarm target we show they are complementary, that fusion covers both, and we quantify the tightened targeted-bias budget.

\subsection{Evaluation protocol}\label{sec:proto}
Two effects can make a time-series detector look better than it is. The first is adjacency leakage. Neighbouring windows are strongly correlated, so a shuffled split places near-copies of a test window in the training set. The second is drift. The data distribution changes over time, so a model trained on early data meets conditions it has not seen. The protocols below are ordered by how much of each effect they remove. On dataset~1 ($1{,}529$ windows) four protocols are run, in increasing order of stringency: shuffled 5-fold cross-validation (CV); time-blocked 5-fold CV, which keeps contiguous time blocks intact; rolling-origin evaluation ($8$ origins; \texttt{exp8\_rolling\_origin.py}); and a chronological hold-out, defined precisely as training on the chronologically first $75\%$ of each class's windows and testing on the final $25\%$, with no shuffling across the boundary. A five-family model comparison is run under both the shuffled and the rolling-origin protocols (\texttt{exp10\_temporal\_baselines.py}). A stratified $75/25$ split is used only for the per-class results in \ref{app:gcn}.

\begin{figure}[t]\centering\includegraphics[width=\linewidth]{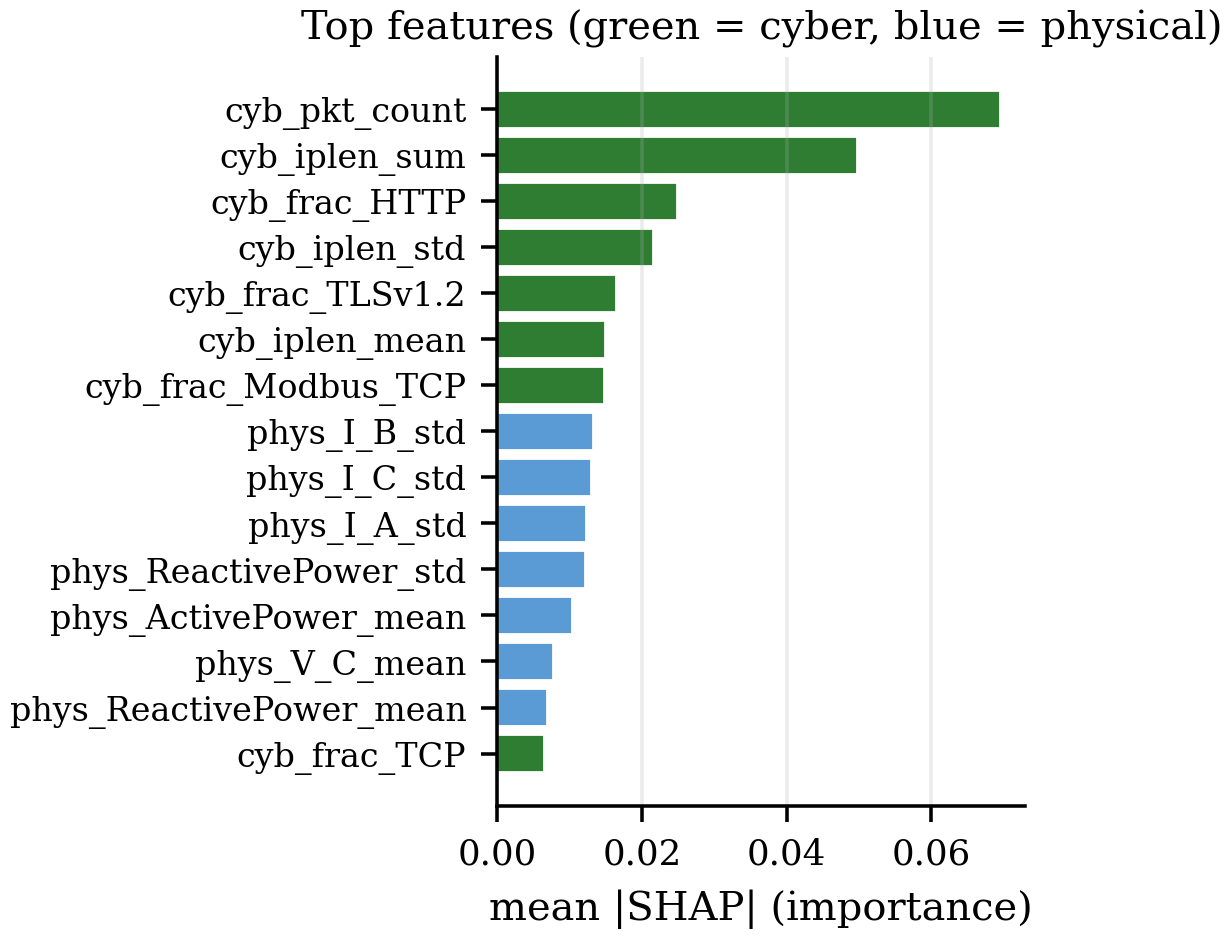}\caption{SHAP importances: both cyber (green) and physical (blue) features matter.}\label{fig:shap}\end{figure}

A further threat to validity comes from how dataset~1 was recorded. Each class is a separate capture, so the class label is confounded with the capture itself. A classifier can therefore separate the classes by learning properties of each recording session rather than properties of the attack. This concern is sharpened by the SHAP ranking (Fig.~\ref{fig:shap}): the two most important features are the packet count and the summed IP length per window, and both could reflect the traffic conditions of a capture as easily as the behaviour of an attack. The temporal protocols do not remove this confound, because every chronological split stays inside each class's own capture. The dataset-1 detection numbers should therefore be read as an upper bound on attack separability, and confirming them requires data in which several classes share one capture session.

\subsection{Threat models and defenses}\label{sec:threat}
\subsubsection{Measurement-space threat model (dataset~1)}
The adversary perturbs the detector's input features within an $L_\infty$ ball; this is an approximation that does not enforce grid physics (no topology available). We grade four attackers: query-based random search; PGD crafted on a differentiable surrogate and transferred~\cite{papernot2016transferability}; a target-aware \emph{adaptive} attack, in which the attacker queries the deployed detector itself rather than a surrogate and adapts perturbations to its decision boundary; and a joint attacker that also perturbs cyber features. Because dataset~1 exposes no topology, these attackers edit the detector's input features directly, so they are feature-space stress bounds: within the same budget they bound the damage of any attack, but they are not themselves feasible grid manipulations, and the joint attacker in particular edits cyber features that a real adversary could influence only indirectly.
\subsubsection{Physics-feasible threat model (testbed~2)}
\begin{figure}[t]\centering\includegraphics[width=\linewidth]{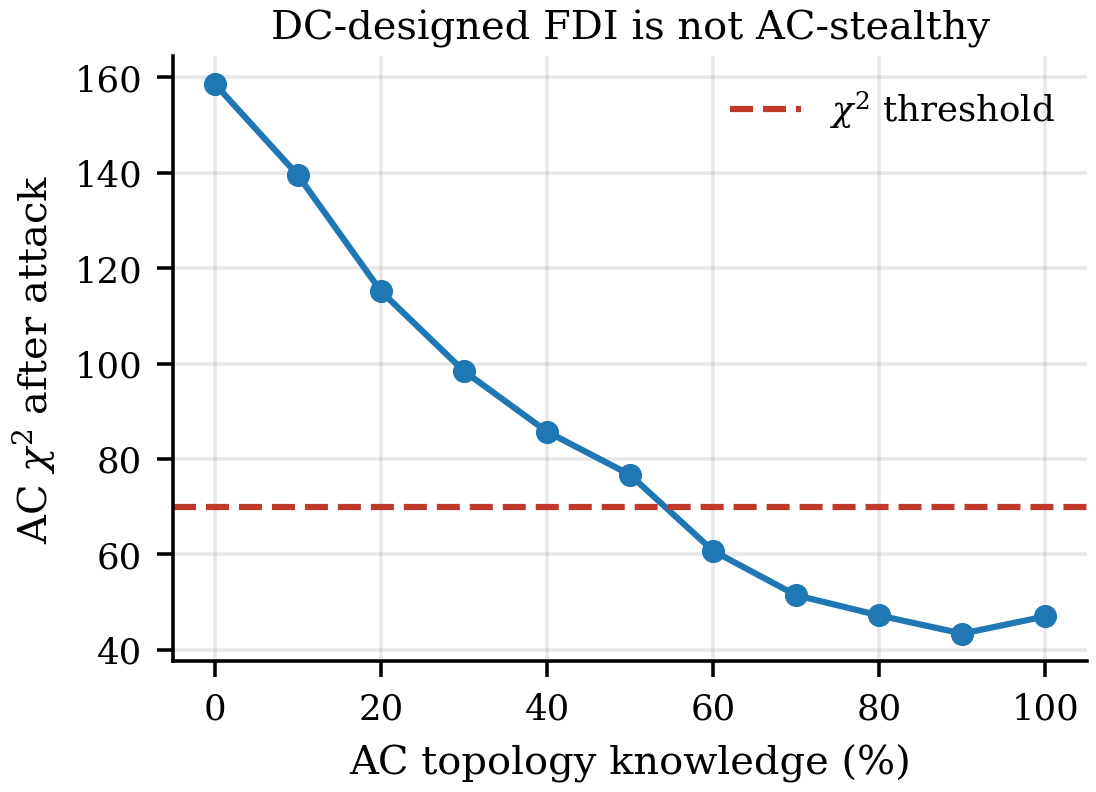}\caption{A DC-designed attacker is not AC-stealthy: the AC chi-square only falls below threshold as the attacker's topology knowledge approaches $50$--$100\%$. Ramp-rate and step-bias detectability are reported in Table~\ref{tab:ramp} and Fig.~\ref{fig:exp6}, where all three detectors are evaluated under a single calibration protocol.}\label{fig:exp5limit}\end{figure}

On IEEE 14-Bus the attacker perturbs the grid state rather than the detector inputs, and a feasible evasion optimizes that state perturbation to fool the detector. We use \emph{physics-feasible} in a precise sense throughout: the attacked measurements are recomputed through the power-flow model of the testbed (the DC model for the feasible-evasion and taxonomy studies, AC power flow for the state-estimation study), so every attacked snapshot is a \emph{valid operating point} of that model; this is distinct from \emph{AC-stealthy}, i.e.\ evading a full nonlinear AC bad-data detector, which a DC-designed injection does \emph{not} in general achieve (Section~\ref{sec:resD4}, Fig.~\ref{fig:exp5limit}). On the IEEE 14-Bus testbed the feasible attacks are provably stealthy \emph{against the DC estimator they are designed for}: because $a=Hc$ lies in the column space of $H$, the DC least-squares residual is unchanged and sits at the metering-noise floor (Eq.~\ref{eq:se}), so DC residual-based bad-data detection cannot flag them (stealthiness against the full AC estimator requires the AC-exact construction of Section~\ref{sec:se}). Physics is therefore enforced only on the IEEE testbeds: on IEEE 14-Bus, and in the full end-to-end run on IEEE 118-Bus.

All attackers solve
\begin{equation}\label{eq:atk}\boldsymbol{\delta}^\star=\arg\min_{\boldsymbol{\delta}\in S}\ f_{\hat y}(\mathbf{x}+\boldsymbol{\delta}),\end{equation}
over the physical-only or joint constraint set
\begin{align}\label{eq:cset}
S_{\text{phys}}&:\ \|\boldsymbol{\delta}_P\|_\infty\le\varepsilon\sigma_P,\quad \delta_{j\notin P}=0;\nonumber\\
S_{\text{joint}}&:\ \|\boldsymbol{\delta}_P\|_\infty\le\varepsilon\sigma_P,\quad \|\boldsymbol{\delta}_C\|_\infty\le\varepsilon\sigma_C.
\end{align}
The projected-gradient-descent (PGD) attacker approximates \eqref{eq:atk} subject to \eqref{eq:cset} on a differentiable surrogate and transfers~\cite{papernot2016transferability}:
\begin{equation}\mathbf{x}^{k+1}=\Pi_{B_\varepsilon(\mathbf{x})}\!\big(\mathbf{x}^{k}+\alpha\,\operatorname{sign}(\nabla_{\mathbf{x}}\mathcal{L}(f_\theta(\mathbf{x}^{k}),y))\big),\end{equation}
where $\Pi_{B_\varepsilon(\mathbf{x})}$ projects onto the $\varepsilon$-ball around $\mathbf{x}$, $\alpha$ is the step size, and $\mathcal{L}$ the loss of the differentiable surrogate $f_\theta$ whose gradients substitute for the forest's.

A feasible evasion optimizes $\mathbf{c}$ ($\|\mathbf{c}\|_\infty\le\gamma$) with a $(1,\lambda)$ evolution strategy:
\begin{align}\label{eq:es}
\mathbf{c}^{(g+1)}&=\operatorname{clip}\!\big(\boldsymbol{\mu}^{(g)}+s^{(g)}\boldsymbol{\xi}\big),\qquad s^{(g+1)}=\beta s^{(g)},\nonumber\\
\boldsymbol{\mu}^{(g+1)}&=\arg\min_{\mathbf{c}\in\text{pop}}\Pr\nolimits_{\text{RF}}[\text{atk}\mid \mathbf{z}{+}\mathbf{H}\mathbf{c}],
\end{align}
where $\boldsymbol{\xi}\sim\mathcal{N}(\mathbf{0},\mathbf{I})$, $s^{(g)}$ is the mutation step size at generation $g$ (denoted $s$ because $\sigma$ is reserved for standard deviations), $\beta<1$ its decay factor, and the arg-min selects the population member the detector misclassifies most confidently.

Two defenses are evaluated: adversarial training, which retrains the classifier on worst-case perturbed windows, and randomized smoothing, which classifies by majority vote under Gaussian input noise~\cite{madry2018pgd,cohen2019certified}:
\begin{align}
&\min_{h}\mathbb{E}\big[\max_{\boldsymbol{\delta}\in S}\mathcal{L}(h(\mathbf{x}{+}\boldsymbol{\delta}),y)\big],\nonumber\\
&g(\mathbf{x})=\arg\max_c\Pr\nolimits_{\mathbf{u}\sim\mathcal{N}(\mathbf{0},\sigma_{\text{sm}}^2\mathbf{I})}[f(\mathbf{x}{+}\mathbf{u}){=}c],
\end{align}
where $\mathbf{u}$ is the smoothing noise, of standard deviation $\sigma_{\text{sm}}$.

\subsection{Classifier and metrics}\label{sec:clf}
An ensemble of $T$ decision trees $h_t$ yields a predictive distribution and a balanced class weighting:
\begin{align}
p_c(\mathbf{x})&=\tfrac{1}{T}\!\sum_{t=1}^{T}\mathbf{1}[h_t(\mathbf{x})=c],\nonumber\\
\hat{y}&=\arg\max\nolimits_c p_c(\mathbf{x}),\qquad w_c=\tfrac{N}{|C|N_c},
\end{align}
where $N$ is the number of training windows, $C$ the class set, $N_c$ the number of training windows of class $c$, and $w_c$ the weight applied to class $c$ during training.
Feature attributions use Shapley values, where $F$ is the feature set, $S\subseteq F$ a feature subset (upper case by game-theoretic convention), and $f(S)$ the model output restricted to $S$~\cite{lundberg2017shap}:
\begin{equation}\phi_i=\!\!\sum_{S\subseteq F\setminus\{i\}}\!\!\tfrac{|S|!(|F|-|S|-1)!}{|F|!}\big(f(S\cup\{i\})-f(S)\big).\end{equation}

From the predictive distribution a confidence and a predictive entropy are derived,
\begin{equation}\kappa(\mathbf{x})=\max_c p_c(\mathbf{x}),\qquad \mathcal{H}(\mathbf{x})=-\!\sum_c p_c(\mathbf{x})\log p_c(\mathbf{x}),\end{equation}
where $\kappa$ is the maximal class probability (a scalar confidence, distinct from the arg-max prediction $\hat y$) and $\mathcal{H}$ is the Shannon entropy of the predictive distribution (calligraphic, as distinct from the Jacobian $\mathbf{H}$),
and define a selective predictor $(f,g)$ that outputs $f(\mathbf{x})$ when $g(\mathbf{x})=\mathbf{1}[\kappa(\mathbf{x})\ge\tau]$ and abstains otherwise, with coverage and selective risk
\begin{equation}\psi(\tau)=\mathbb{E}\big[\mathbf{1}(\kappa\ge\tau)\big],\qquad R(\tau)=\frac{\mathbb{E}\big[\mathcal{L}(f(\mathbf{x}),y)\,\mathbf{1}(\kappa\ge\tau)\big]}{\psi(\tau)},\end{equation}
with $\mathcal{L}$ the classification loss. Abstained windows are escalated to physical-layer verification, operationalizing fusion as a decision policy. The abstention threshold $\tau$ can be set operationally by trading off coverage against false-positive rate; in practice, a grid operator might choose the knee of the risk--coverage curve.

Reported metrics are per-class precision and recall and the macro-averaged F1 (macro-F1), the unweighted mean of per-class F1 scores. Macro-F1 is used as a single summary of multi-class performance, but the operationally meaningful quantities for protection and security practice are the \emph{per-attack recall}, the \emph{false-alarm rate} at the chosen operating point, and, for the temporal detectors, the \emph{detection delay}; these are reported alongside macro-F1 wherever they are available and should be read as the primary indicators:
\begin{equation}P_c=\tfrac{TP_c}{TP_c+FP_c},\ R_c=\tfrac{TP_c}{TP_c+FN_c},\ \text{macro-F1}=\tfrac{1}{|C|}\!\sum_c\tfrac{2P_cR_c}{P_c+R_c}.\end{equation}

\section{Test Setup and Scenarios}\label{sec:setup}
The detector is a Random Forest ($300$ trees, balanced class weights)~\cite{breiman2001random,pedregosa2011scikit}; the PGD surrogate is a two-layer MLP; the attack budget is $\varepsilon=0.10$ and the smoothing noise $\sigma_{\text{sm}}=0.10$. Window widths are matched to each stream's event rate: dataset-1 attacks manifest as sub-second bursts in high-rate captures spanning minutes, so $\Delta=0.5$\,s; IEC-104 HMI traffic is slow (periodic polling and spontaneous events), so $120$\,s windows are required to accumulate a statistically meaningful packet count per window. All experiments fix random seeds and are regenerated end-to-end by the released scripts (Section~\ref{sec:conc}).

\emph{Protocol ladder for the Random Forest (Section~\ref{sec:resD1}).} The RF alone ($300$ trees, balanced weights) is run through four protocols of increasing stringency: shuffled CV, time-blocked CV, rolling-origin, and a strict $75/25$ chronological hold-out ($n=381$). Physical-only and cyber-only ablations were run under shuffled CV only.

\emph{Protocol comparison across model families (Section~\ref{sec:resD1}).} Five model families (RF, histogram gradient boosting, per-window MLP, TDNN over $5$ windows, logistic) are trained on the same $1{,}529$ fused $0.5$\,s windows and evaluated twice: shuffled 5-fold CV, then rolling-origin with $8$ origins from $50\%$ to $85\%$ of each class timeline ($n\approx230$ per block). Only the protocol varies.

\emph{Feasible evasion on IEEE 14-Bus (Section~\ref{sec:resD2}).} On an IEEE 14-Bus \texttt{pandapower} DC testbed, $1{,}200$ benign and $1{,}200$ feasible $\mathbf{a}=\mathbf{H}\mathbf{c}$ states train an RF at a $70/30$ split (stability checked over $15$ splits), then a $(1,\lambda)$ evolution strategy runs $8$ generations over the state offset $\mathbf{c}$ to minimize the detector's attack probability, also under a limited-observability variant with unmetered branches.

\emph{Feasible-space adversarial training (Section~\ref{sec:resD2}).} The same testbed is rerun with a training set augmented by feasible $\mathbf{a}=\mathbf{H}\mathbf{c}$ perturbations, including ES-mined evasions, comparing clean performance and recall under ES evasion against the undefended detector.

\emph{Real IEC-104 traffic (Section~\ref{sec:resD3}).} Real IEC-104~\cite{iec60870} HMI captures~\cite{matousek2022ics} are cut into $7{,}107$ windows of $120$\,s ($637$ attack, $6{,}470$ benign), labeled from documented attack intervals, and scored by 5-fold out-of-fold detection on cyber features only, with Wilson intervals per attack class at $2.4\%$ FPR.

\emph{Observability taxonomy on IEEE 14-Bus (Section~\ref{sec:resD3}).} On the same DC model (Eq.~\ref{eq:se}, Jacobian $\mathbf{H}$) we generate benign operating points and three attack classes: (A)~a \emph{structural} attack that perturbs a simulated cyber/message-feature vector while leaving the measurements physically consistent; (B)~a \emph{naive value} attack in which an operator- or MITM-level adversary alters reported measurement values \emph{without topology knowledge}, so the perturbation leaves the column space of $\mathbf{H}$ and inflates the state-estimation residual; and (C)~the \emph{stealthy FDI} $\mathbf{a}=\mathbf{H}\mathbf{c}$ of Section~\ref{sec:resD2}, whose residual stays at the metering-noise floor. A single detector is trained on a mixture of all three (it is not told the class) under three feature views: cyber features only, the physics residual vector $\mathbf{r}=\mathbf{z}-\mathbf{H}\hat{\mathbf{x}}$ only, and their fusion. Every detector's decision threshold is calibrated on held-out benign windows to a common $5\%$ false-positive rate, and we then read per-class recall at that operating point (Table~\ref{tab:tax}).

\emph{Command-transition model (Section~\ref{sec:resD3}).} Per $120$\,s window, IEC-104 command symbols (asduType, CoT) are scored against the benign empirical distribution for surprise, novelty and entropy (Eq.~\ref{eq:cmd}), testing whether any header-level sequence statistic separates the semantic attacks.

\emph{Ramp sweep on the IEEE 14-Bus AC testbed (Section~\ref{sec:resD4}).} A full AC WLS estimator on IEEE 14-Bus ($72$ measurements, $27$ states) runs over time-varying loads. An AC-exact stealthy FDI is ramped from $0.50$ to $0.02^\circ$/step ($50$ trials per rate), against a Holt/CUSUM temporal test, a PCA off-manifold test fitted on estimated states with $K{=}6$ selected by cross-validation on benign data alone, and their normalized maximum, all calibrated to a $5\%$ false-alarm rate.

\emph{End-to-end IEEE 118-Bus evaluation (Section~\ref{sec:resD4}).} We finally test whether the picture survives beyond IEEE 14-Bus with a \emph{full end-to-end} evaluation on IEEE 118-Bus ($235$ states, $700$ measurements, redundancy $2.98$). Nothing is extrapolated: the AC weighted least-squares estimator is solved at every timestep, with the Jacobian recomputed at every Gauss--Newton iteration, and the detectors read the resulting estimated-state stream. The study uses $140$ benign streams for calibration, $90$ independent benign streams to measure the achieved false-alarm rate, and $50$ attacked trials at each ramp rate, each stream $40$ steps long (\texttt{exp7c\_ieee118\_full.py}). The manifold dimension is selected by cross-validation on the benign streams alone, with no access to attacked data; this yields $K{=}8$. A cumulative-variance rule is not used here either. The estimated states of the larger network carry substantially more estimation noise, the benign spectrum has no clear low-rank gap, and a $99.9\%$-variance rule would select $K{=}165$, producing a badly miscalibrated detector. Selecting $K$ by benign cross-validation restores calibration, with achieved false-alarm rates of $0.067$, $0.011$ and $0.056$ for the temporal, manifold and fused detectors against a $5\%$ target.

\emph{Bias budget and geometric bound (Section~\ref{sec:resD4}).} The off-manifold residual is swept against injected targeted bias from $0$ to $3^\circ$ on the IEEE 14-Bus AC testbed to locate the detection gate at three benign standard deviations of the residual ($3\sigma_\rho$), complemented by a geometric bound on how much of a single-coordinate bias is forced off the manifold (\texttt{exp9\_manifold\_attacker.py}).

\section{Results}\label{sec:results}
\subsection{Evaluation protocols on dataset~1}\label{sec:resD1}
This section supports Claim~1. It shows that the gap between shuffled and honest evaluation is a property of the protocol and not of the interpretable detector.

\begin{figure}[t]\centering\includegraphics[width=\linewidth]{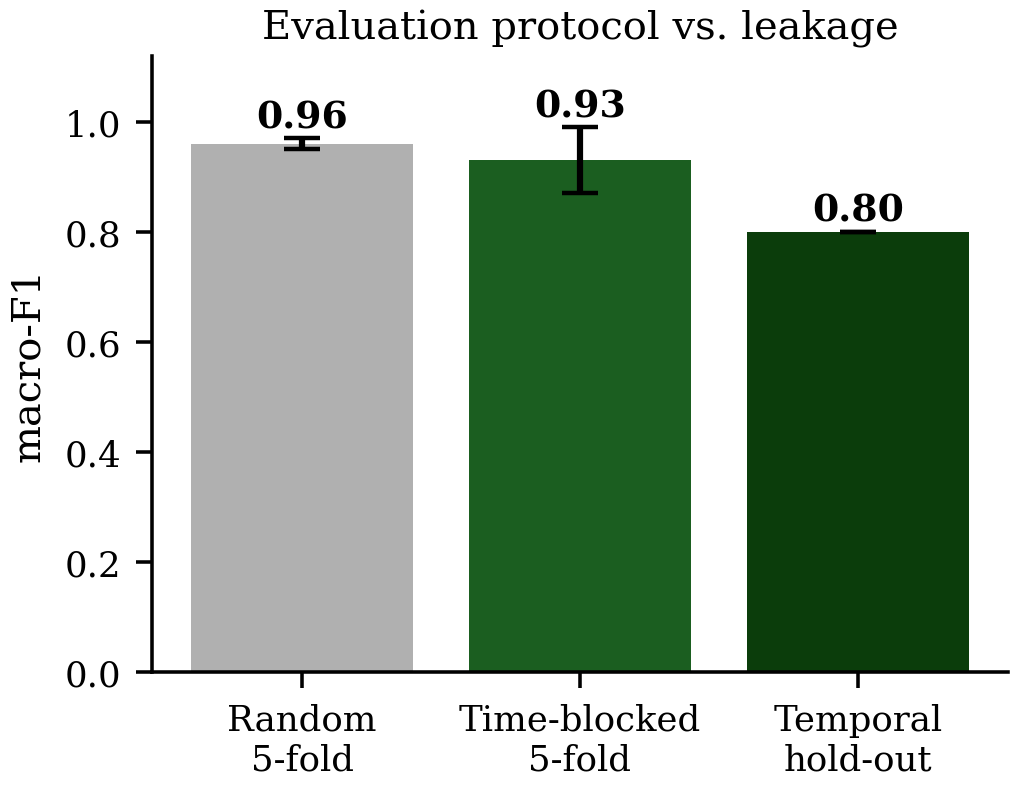}\caption{Evaluation protocol vs.\ leakage: shuffled CV is optimistic; time-blocked CV removes adjacency leakage; a temporal hold-out exposes distribution shift.}\label{fig:blk}\end{figure}

\begin{figure}[t]\centering\includegraphics[width=\linewidth]{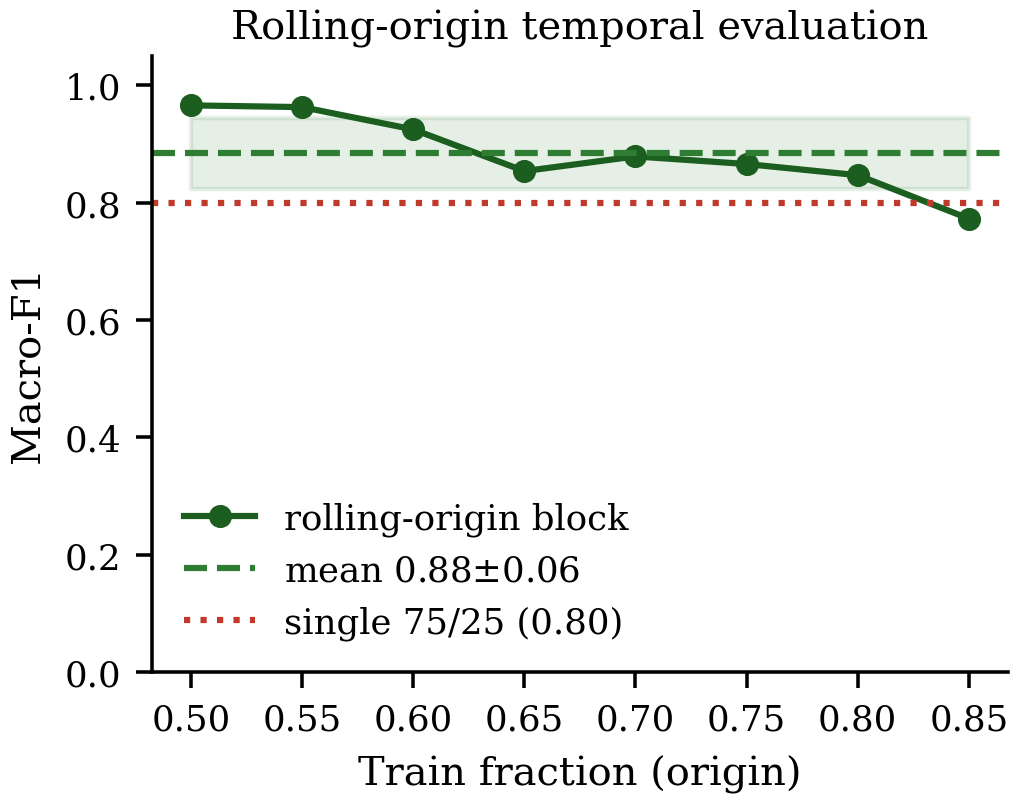}\caption{Rolling-origin temporal evaluation: eight strictly-temporal train/test splits give $0.88\pm0.06$ macro-F1, declining for later origins (drift); the single $75/25$ split ($0.80$) sits at the pessimistic end.}\label{fig:ro}\end{figure}

Fusion beats either single layer and every class reaches at least $0.93$ F1 (Tables~\ref{tab:det} and~\ref{tab:perclass}). We evaluate under the four protocols of Section~\ref{sec:proto} (Fig.~\ref{fig:blk}). Shuffled 5-fold CV can place temporally adjacent (hence correlated) windows in both train and test; \emph{time-blocked} 5-fold CV, which keeps contiguous time blocks intact, removes this leakage and lowers the score modestly to $0.93\pm0.06$ (a $0.03$ leakage gap). A strict temporal hold-out (train on the first $75\%$ in time, test on the last $25\%$, $n{=}381$ test windows) drops further to $0.80$: distribution shift over time, not leakage, is the dominant evaluation cost. Because a single split at this sample size (roughly $50$ to $90$ test windows per class) is a point estimate with no sampling variability, we additionally run \emph{rolling-origin} evaluation: for eight origins from $50\%$ to $85\%$ of each class's timeline, we train on everything before the origin and test on the next contiguous $15\%$ block ($n{\approx}230$ per block). The rolling-origin macro-F1 is $0.88\pm0.06$ (range $0.77$--$0.97$; Fig.~\ref{fig:ro}), declining overall for later origins (direct evidence of drift, with minor fluctuation between the $0.65$ and $0.75$ origins), with the paper's single $75/25$ split ($0.80$) sitting at the pessimistic end of that distribution. We therefore report the temporal performance as $0.88\pm0.06$ with a worst-block value of $0.77$, and treat $0.80$ as a representative pessimistic operating point rather than a precise headline.

\begin{table}[t]\centering\caption{Dataset-1 detection (macro-F1). The GCN is a topology-free control built for this study, not the topology-aware fusion GNN of Sweeten et al.~\cite{sweeten2025fusion}.}\label{tab:det}
\footnotesize\setlength{\tabcolsep}{4pt}
\begin{tabular}{@{}p{0.62\linewidth}c@{}}\toprule Setting & macro-F1\\\midrule
Physical-only (shuffled CV) & $0.77\pm0.02$\\
Cyber-only (shuffled CV) & $0.85\pm0.01$\\
Fusion (shuffled 5-fold CV, optimistic) & $0.96\pm0.01$\\
Fusion (time-blocked 5-fold CV) & $0.93\pm0.06$\\
\textbf{Fusion (rolling-origin temporal)} & $\mathbf{0.88\pm0.06}$\\
Fusion (final temporal hold-out, $n{=}381$) & $0.80$\\
GCN control, topology-free (shuffled / temporal) & $0.83$ / $0.70$\\
Gradient boosting / MLP (shuffled CV) & $0.95$ / $0.91$\\\bottomrule\end{tabular}\end{table}


\begin{figure*}[t]\centering\includegraphics[width=\textwidth]{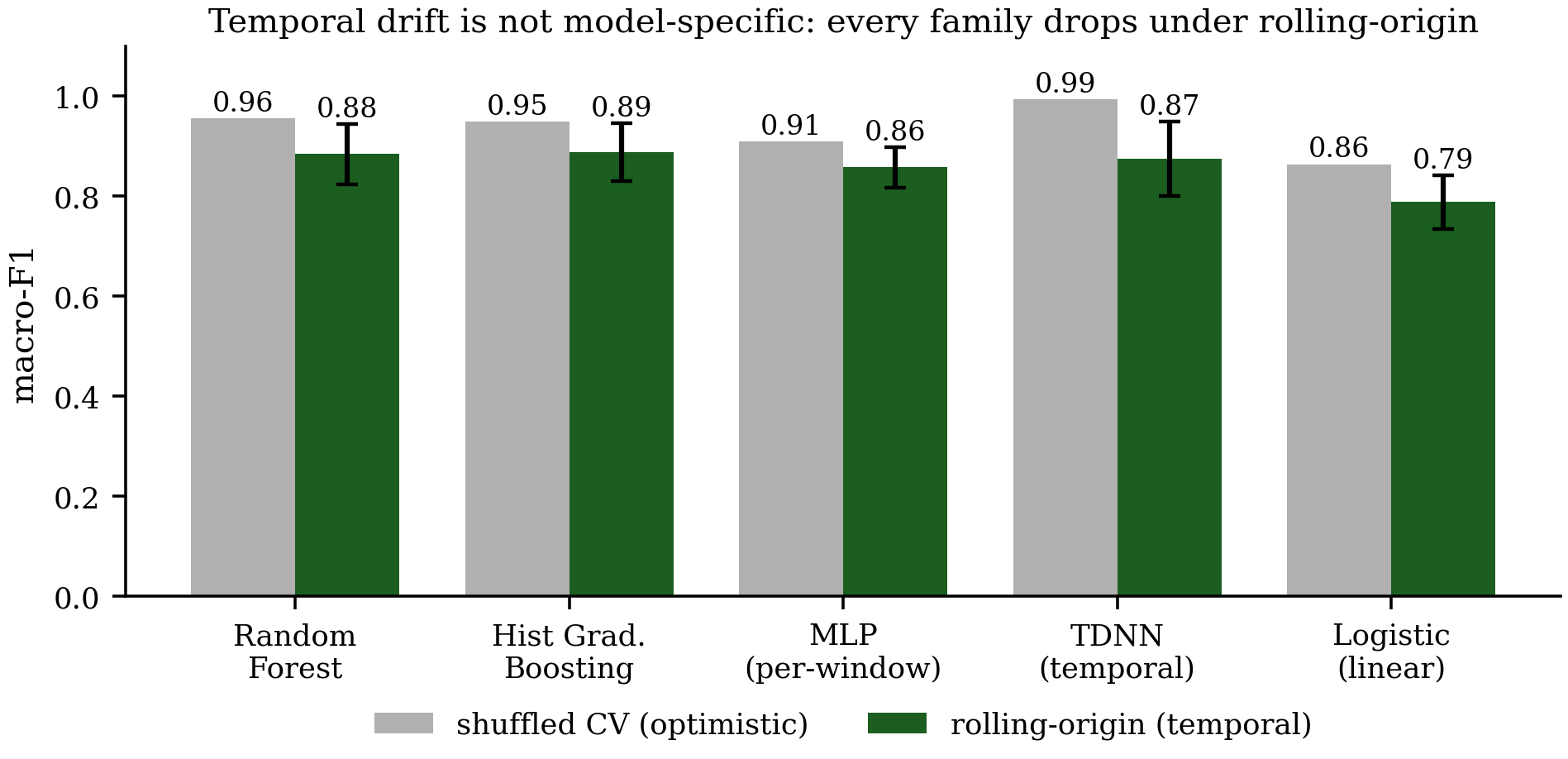}\caption{The temporal drop is not model-specific: five model families, including a temporal time-delay neural network, all fall from shuffled CV to rolling-origin evaluation. The TDNN is the most inflated by shuffling and the most penalized by honest evaluation.}\label{fig:tbase}\end{figure*}

This drift is a property of the data, not of the Random Forest. Under identical shuffled and rolling-origin protocols, four additional model families all fall by a comparable margin (Fig.~\ref{fig:tbase}): histogram gradient boosting ($0.95\!\to\!0.89$), a per-window MLP ($0.91\!\to\!0.86$), a temporal time-delay neural network (TDNN, an MLP over the current window and its four predecessors; $0.99\!\to\!0.87$), and a linear logistic baseline ($0.86\!\to\!0.79$). Tellingly, the TDNN, the model that most exploits temporal structure, is both the most flattered by shuffled CV ($0.99$, the highest of any model) and the hardest hit by honest evaluation (a $0.12$ drop, the largest observed): a heavier temporal model does not escape distribution shift but is if anything more exposed to it. The gap between protocols is therefore not an artifact of the interpretable detector; it is intrinsic to chronologically evaluating this benchmark.

A recurrent LSTM under the identical protocol is released as a script (\texttt{exp10b\_lstm\_optional.py}) for reproduction where a deep-learning backend is available; it is not part of the comparison above.

\subsection{Feature-space and physics-feasible attacks}\label{sec:resD2}
This section also supports Claim~1. It shows that confining the attacker to physically feasible states does not rescue the detector.

\begin{figure}[t]\centering\includegraphics[width=\linewidth]{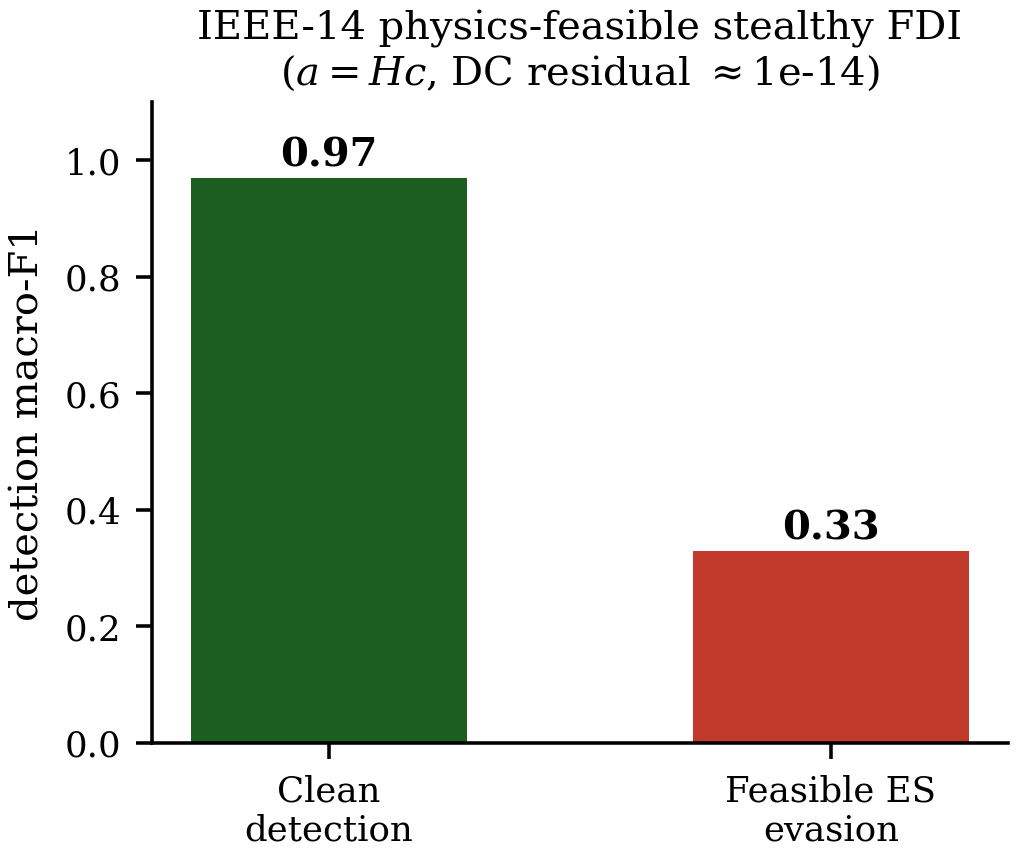}\caption{Physics-feasible stealthy FDI on IEEE 14-Bus ($a=Hc$; DC residual unchanged at the metering-noise floor); an ES evasion defeats the detector. Macro-F1: $0.97$ clean, $0.33$ under evasion; the limited-observability recall analysis is in the text.}\label{fig:ieee}\end{figure}

\begin{figure}[t]\centering\includegraphics[width=\linewidth]{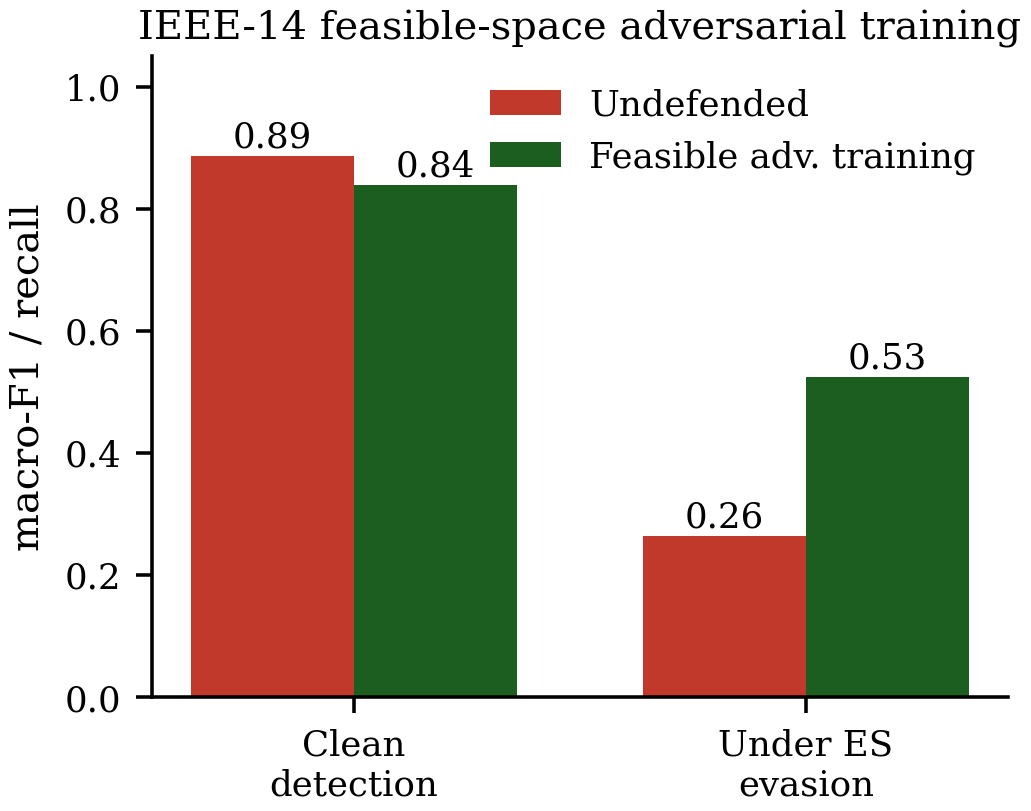}\caption{IEEE 14-Bus: a feasible ES evasion collapses undefended detection; feasible-space adversarial training partially recovers recall on the retained set.}\label{fig:adv}\end{figure}

To keep metrics distinct: Fig.~\ref{fig:ieee} reports macro-F1, where a clean detector scores $0.97$ and the ES evasion (Eq.~\ref{eq:es}) drives it to $0.33$. Because the operationally meaningful quantity is how many injected windows are missed, the remaining discussion reports the FDI-class \emph{recall}. Under \emph{limited observability}, the classical FDI setting where the adversary exploits unmetered branches~\cite{liu2011fdi}, the ES evasion, which fixes a meaningful target-state bias and shapes the remaining feasible freedom, collapses that recall to $0.26$ for a detector whose clean macro-F1 is $0.89$ (Fig.~\ref{fig:adv}). Across $15$ repeated stratified $70/30$ splits the clean feasible-FDI detector is stable at macro-F1 $0.951$ ($95\%$ CI $[0.947,0.954]$, range $0.935$--$0.960$), so the single-split figures quoted here are representative rather than lucky draws; the evasion collapse is a security finding, not resampling noise. Feasible-space adversarial training, augmenting the training set with feasible $a=Hc$ perturbations including ES-mined evasions, recovers the recall to $0.53$ at a small clean-accuracy cost (clean macro-F1 $0.84$), materially better than naive adversarial training, though it does not fully close the gap. This motivates certified and moving-target defenses~\cite{mtd2025,cohen2019certified} as priorities.

\begin{figure}[t]\centering\includegraphics[width=\linewidth]{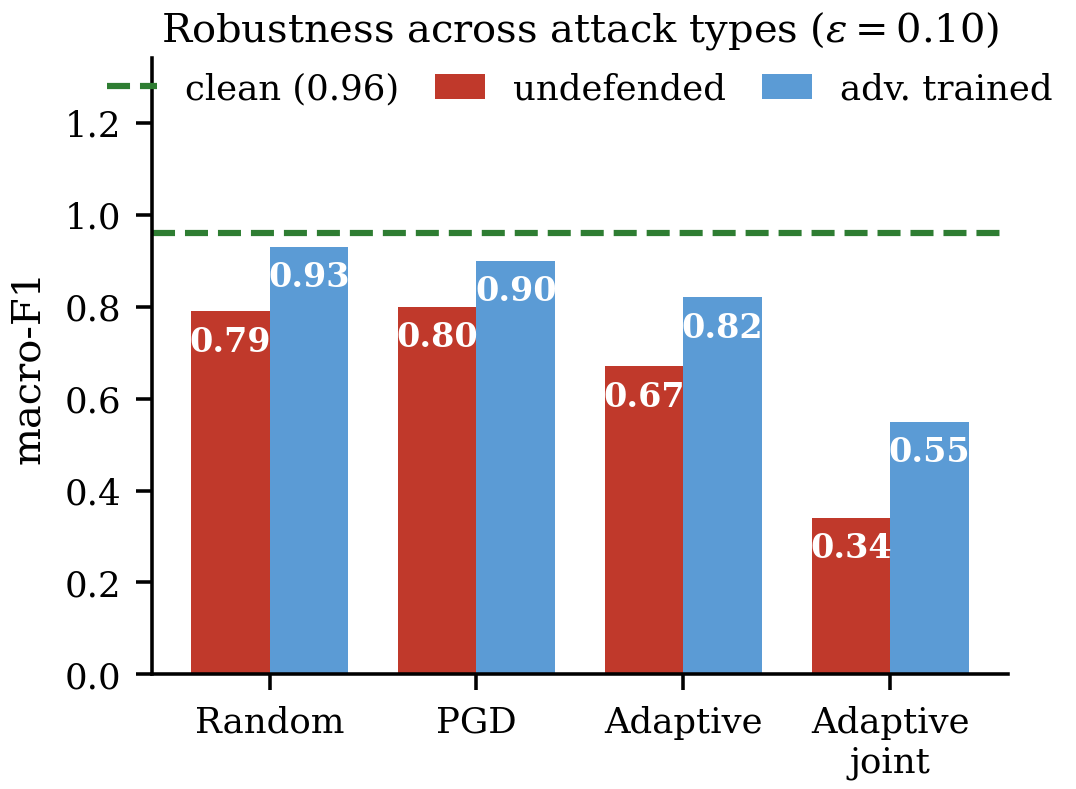}\caption{Dataset-1 robustness across attack types ($\varepsilon{=}0.10$); adversarial training retains a margin.}\label{fig:spec}\end{figure}

Robustness is graded (Table~\ref{tab:rob}, Fig.~\ref{fig:spec}): the joint attack ($0.34$) shows the physical-only threat model is optimistic. Randomized smoothing yields a per-input certified radius, not a certified F1~\cite{cohen2019certified}; what we report is the \emph{empirical} macro-F1 of the smoothed classifier under the adaptive attack, $0.80$. This robustness is bought at a real clean-performance cost: at $\sigma_{\text{sm}}{=}0.10$ the measured benign false-positive rate rises to $9.1\%$ (an outcome measured on held-out benign windows, not a design choice), roughly one alarm per eleven benign windows, which would need to be weighed against analyst workload before deployment. Training takes $0.8$\,s and inference $0.30$\,ms/window.

\begin{table}[t]\centering\caption{Dataset-1 robustness (macro-F1, $\varepsilon{=}0.10$). Clean $=0.96$. These are \emph{feature-space} worst cases (attacker edits detector inputs directly; no grid feasibility enforced), so they bound rather than represent realizable attacks.}\label{tab:rob}
\footnotesize\begin{tabular}{lcc}\toprule Attack & Undefended & Adv.\ trained\\\midrule
Random search & $0.79\pm0.01$ & 0.93\\
PGD transfer & 0.80 & 0.90\\
Adaptive & 0.67 & 0.82\\
Adaptive joint & \textbf{0.34} & 0.55\\\bottomrule\end{tabular}\end{table}

\subsection{The detection boundary of the cyber view}\label{sec:resD3}
This section supports Claim~2. It shows that what a detector can see is set by the layer it observes.

\begin{figure}[t]\centering\includegraphics[width=\linewidth]{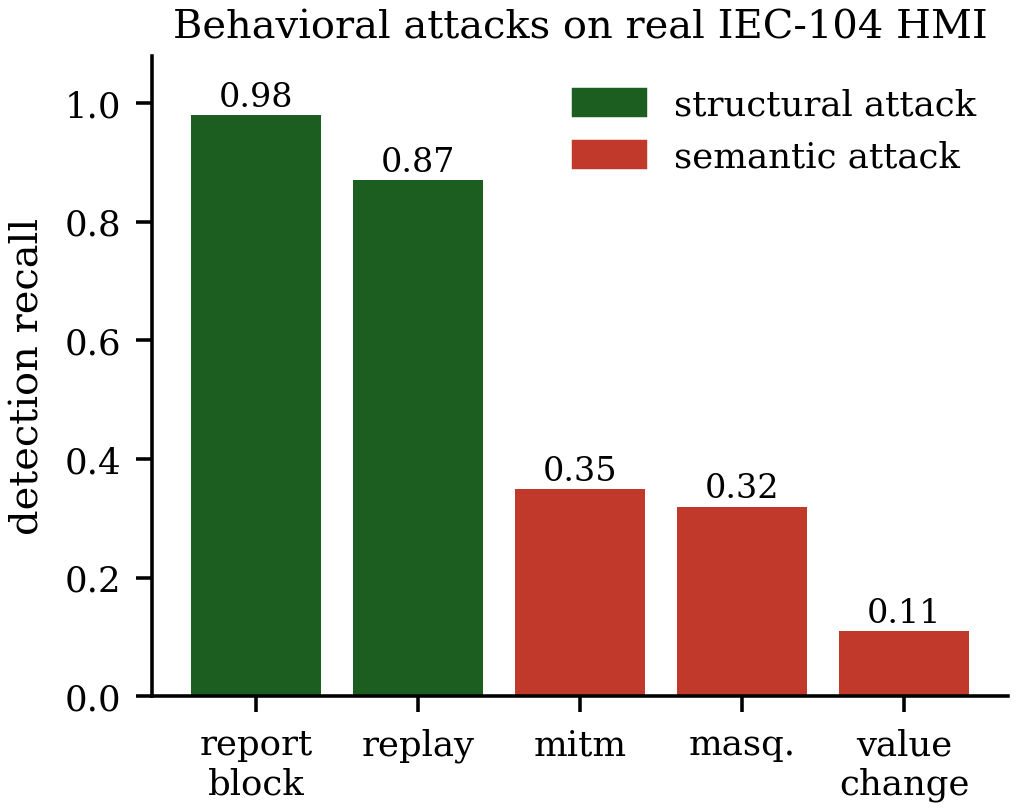}\caption{Behavioral attacks on real IEC-104 HMI traffic ($637$ attack windows); structural attacks (green) are caught, while semantic attacks (red) evade the cyber layer.}\label{fig:hmi}\end{figure}

On real IEC-104 HMI captures with five behavioral attacks (Table~\ref{tab:iec}, Fig.~\ref{fig:hmi}), structural attacks are caught (report-block $0.98$, replay $0.87$) at a $2.4\%$ false-positive rate (at the $120$\,s IEC-104 window rate, about six false alarms per $8$-hour shift). We report this figure descriptively and make no claim of operational acceptability: sustained nuisance alarms at this rate would risk alarm fatigue in a real control room, so a deployable configuration would require a substantially lower false-alarm rate than the research operating point used here, but semantic value-manipulation attacks largely evade the network view (MITM $0.35$, masquerade $0.32$, value-change $0.11$).

\begin{table}[t]\centering\caption{Dataset-3 per-attack detection recall on real IEC-104 HMI traffic, with Wilson $95\%$ confidence intervals over the attack windows of each class ($5$-fold out-of-fold; benign windows $n{=}6470$, FPR $0.024$). The structural and semantic classes separate with non-overlapping intervals.}\label{tab:iec}
\footnotesize\begin{tabular}{lccc}\toprule Attack & Type & Recall (95\% CI) & Windows\\\midrule
report-block & structural & 0.98 [0.92, 1.00] & 65\\
replay & structural & 0.87 [0.79, 0.92] & 105\\
MITM & semantic & 0.35 [0.29, 0.42] & 207\\
masquerade & semantic & 0.32 [0.25, 0.40] & 148\\
value-change & semantic & 0.11 [0.06, 0.18] & 112\\\bottomrule\end{tabular}\end{table}

The IEC-104 results above establish, on real traffic, that semantic value-manipulation is invisible to the cyber view. That finding raises a sharper question than ``does fusion help?'': \emph{which} observation layer is required to see \emph{which} attack, and where does even fusion fail? We answer it on the physics-controlled IEEE 14-Bus testbed, where, unlike Dataset~3, we command the ground-truth grid state and can therefore attribute detectability to the observable rather than to a particular capture. We stress that the semantic manipulation studied here is not present in the header-only features of Dataset~3; the value it corrupts lives in the measurement payload, which the IEEE 14-Bus testbed lets us model explicitly.

\begin{table}[t]\centering\caption{IEEE 14-Bus detection recall per attack class at a calibrated $5\%$ benign false-positive rate, averaged over $20$ independent seeds with Wilson $95\%$ confidence intervals on the pooled test windows (${\sim}9{,}100$ benign windows). LS residuals: benign $3.9{\times}10^{-3}$, naive-value $6.8{\times}10^{-2}$, stealthy-FDI $3.9{\times}10^{-3}$. Pooled over seeds the calibrated benign FPR settles at $0.059$--$0.062$; the $0.08$ seen on any single split is finite-sample calibration noise, not a systematic bias. All non-bold recalls have $95\%$ CIs within $\pm0.02$; bold cells are $\ge0.99$ with CI upper bound $1.00$.}\label{tab:tax}
\footnotesize\setlength{\tabcolsep}{4pt}\begin{tabular}{lcccc}\toprule
View & structural & naive value & stealthy FDI & FPR\\\midrule
Cyber   & \textbf{1.00} & 0.06 & 0.06 & 0.059\\
Physics & 0.06 & \textbf{0.99} & 0.07 & 0.062\\
\textbf{Fusion}  & \textbf{1.00} & \textbf{1.00} & 0.06 & 0.062\\\bottomrule\end{tabular}\end{table}

\begin{figure*}[t]\centering\includegraphics[width=\textwidth]{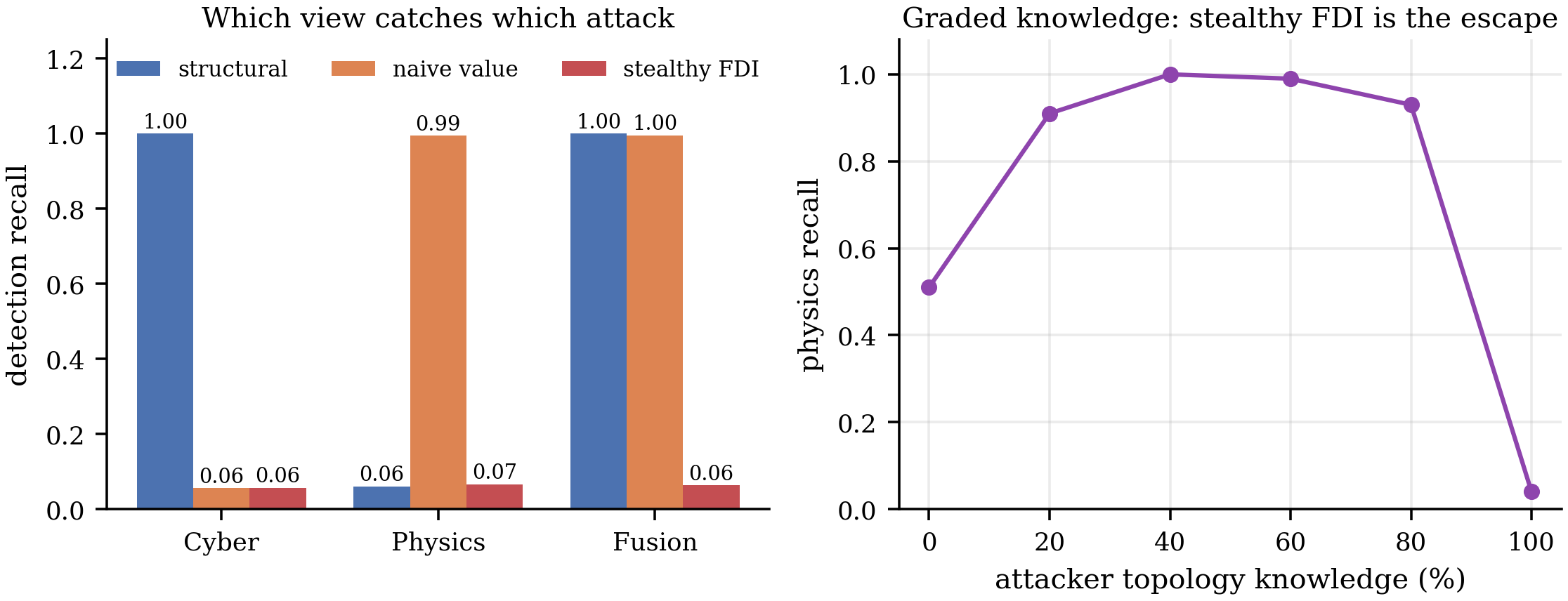}\caption{Left: which view catches which attack (recall pooled over the $20$ seeds of Table~\ref{tab:tax}): cyber sees structural, physics sees naive value manipulation, fusion sees both, and stealthy FDI defeats all three. Right: as the attacker's topology knowledge grows, a partial-knowledge value attack is in fact the \emph{most} detectable (large, inconsistent residual); only at full knowledge does $\mathbf{a}=\mathbf{H}\mathbf{c}$ become consistent and the residual collapse to the noise floor, driving physics recall to the false-positive rate.}\label{fig:tax}\end{figure*}

The three views partition the attack space cleanly. Cyber features detect the structural attack ($1.00$) but are at chance ($\approx$ the $5\%$ FPR) on both value attacks, reproducing the Dataset~3 boundary in a controlled setting (with the caveat that the IEEE 14-Bus cyber features are simulated, so this is a consistency check under a controlled ground truth, not an independent empirical confirmation). The physics residual detects the naive value attack ($0.99$) but is blind to the structural attack (whose measurements are consistent) and to stealthy FDI (whose residual is $\approx$ the noise floor). Fusion inherits the union, structural \emph{and} naive value manipulation, while stealthy FDI remains undetectable to every view (recall $0.06$, i.e.\ at the calibrated false-alarm rate), the fundamental limit of Section~\ref{sec:resD2}; these are exactly the windows the selective policy (\ref{app:sel}) abstains on and escalates.

Mirroring the graded cyber adversary of Section~\ref{sec:resD2}, we sweep the fraction of the injection the adversary can render consistent, from a naive single-branch change to full stealthy FDI (Fig.~\ref{fig:tax}, right). Detectability is \emph{non-monotonic}: a partial-knowledge adversary injects large but inconsistent energy and is the most exposed, whereas only full topology knowledge yields a consistent $\mathbf{a}=\mathbf{H}\mathbf{c}$ that drives the residual to the noise floor and evades the physics view entirely.

We tested the command-transition model (Eq.~\ref{eq:cmd}). The novelty $\nu(w)$ is $\approx 0$ for every attack (they introduce no new command symbols), and the surprise $\zeta(w)$ is elevated only for MITM ($3.66$ vs $2.89$), leaving semantic attacks near chance. This establishes a principled boundary: the manipulated value lives in the payload/physical effect, not the network headers, so no method restricted to these network-layer observables can recover it: the manipulated value is simply absent from the features such a detector reads. The boundary is expected to be protocol-independent on the same grounds (the manipulated value is absent from headers in any SCADA protocol), but a quantitative cross-protocol evaluation on IEC-61850/MMS traffic is not performed here and is left to future work.

That an unbiased state-estimation residual flags inconsistent bad data but not a stealthy $\mathbf{a}=\mathbf{H}\mathbf{c}$ is the classical result of residual-based bad-data detection and of Liu et al.~\cite{liu2011fdi}; we claim no new detector here. The contribution is the \emph{unification}: a single interpretable detector and one physics-controlled testbed that jointly localize which observation layer is necessary and sufficient for each attack class and quantify the residual gap between naive and stealthy manipulation. The cyber layer on IEEE 14-Bus is simulated, as elsewhere in this study.

\subsection{Temporal and spatial detection under AC state estimation}\label{sec:resD4}
This section also supports Claim~2. It shows that the temporal and spatial tests fail in different regimes, and it bounds where the manifold test can be relied upon.

Here we harden the physics testbed to a full AC WLS estimator over time-varying operating points (Section~\ref{sec:method}, Eq.~\ref{eq:wls}) and ask a sharper question: once snapshot bad-data detection is blind to a stealthy FDI, what \emph{can} see it, and where is the boundary? This section directly takes up the stated future work of Sweeten et al.~\cite{sweeten2025fusion}, namely evaluating detection against slow injection attacks that gradually alter measurements over time to mimic stealthy behavior, and characterizes exactly that regime: which detector family a slow ramp evades, why, and what closes the evasion.

Static detection is provably blind. A Jacobian-consistent stealthy FDI (Listing~\ref{lst:fdi}) leaves the estimator's chi-square objective and every normalized residual unchanged: over matched benign/attacked snapshots the two chi-square distributions are statistically identical (mean $J=43.6$ either way; KS $p\approx1.00$; AUC$=0.51$, a coin flip) while the attacker still forces a $4^\circ$ bias onto the target state (Table~\ref{tab:acbdd}); the success criterion is precisely this, a targeted state bias sustained with zero bad-data flags, and the AUC of $0.51$ is the intended demonstration that the snapshot estimator is blind, which motivates the temporal and manifold detectors below. This is the AC analogue of Section~\ref{sec:resD2}, now on the estimator that operators actually run.

\begin{table}[t]\centering\caption{AC WLS state estimation: a Jacobian-consistent stealthy FDI is statistically invisible to snapshot bad-data detection. The $4^\circ$ target bias is achieved with zero bad-data flags.}\label{tab:acbdd}
\footnotesize\setlength{\tabcolsep}{4pt}
\begin{tabular}{@{}lccl@{}}\toprule
Quantity & Benign & Attacked & Verdict\\\midrule
Chi-square $J$ & 43.6 & 43.6 & identical\\
Max LNR & 2.47 & 2.47 & $<3\sigma$\\
Flag rate & n/a & 0\% & undetected\\
Target bias & $0^\circ$ & $4.0^\circ$ & bias achieved\\
Discriminability & \multicolumn{3}{c}{KS test $p{=}1.00$; ROC AUC $=0.51$}\\\bottomrule\end{tabular}\end{table}

\begin{figure*}[t]\centering\includegraphics[width=\textwidth]{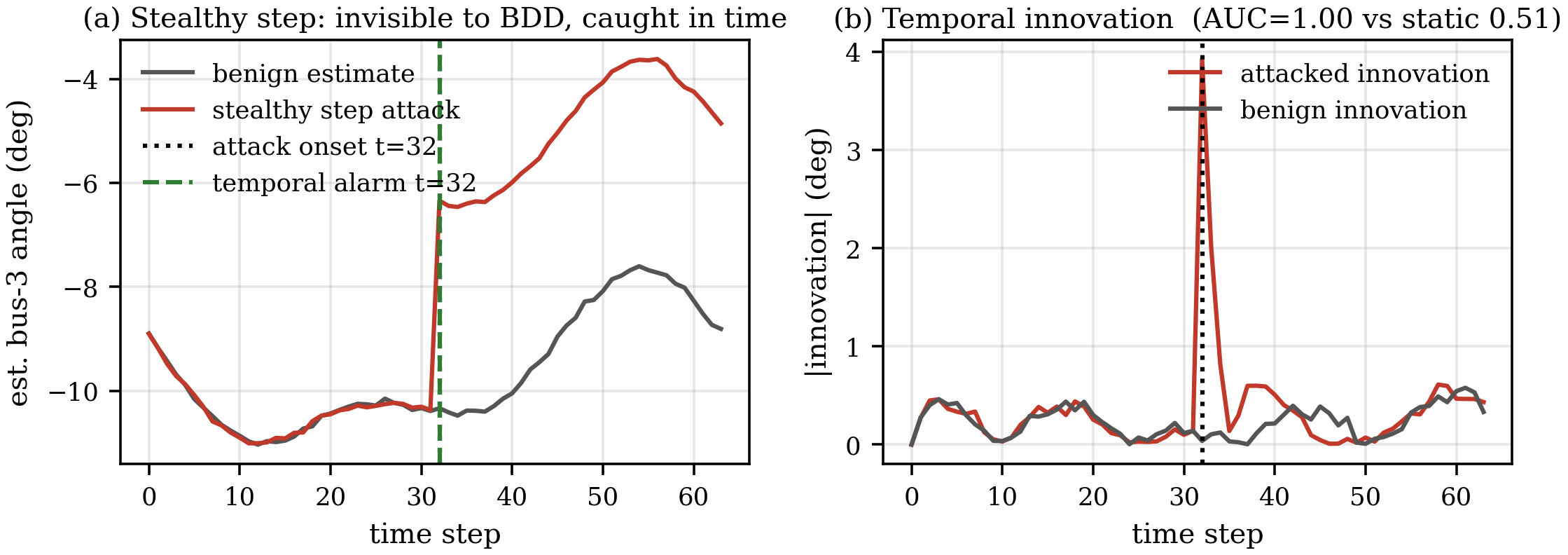}\caption{Real-time AC estimation. A persistent stealthy step at $t{=}32$ shifts the estimated bus-3 angle (left) with the chi-square residual unchanged; the temporal innovation (right) spikes at onset, giving AUC $1.00$ vs.\ $0.51$ for the static test.}\label{fig:exp5temporal}\end{figure*}

A temporal test catches a step but is evaded by a slow ramp. Monitoring the estimated-state stream with a forecasting-aided innovation CUSUM (calibrated to $5\%$ false alarms) detects a persistent stealthy \emph{step} essentially at onset (AUC$=1.00$ vs the static $0.51$; Fig.~\ref{fig:exp5temporal}). But detection is not rate-free: as the injection is ramped in more slowly, the trend filter adapts to it, the innovation returns to the noise floor, and the detection probability falls away, reaching $0.60$ at $0.05^\circ$/step and $0.10$ at $0.02^\circ$/step, where the attack largely escapes the monitoring horizon (Table~\ref{tab:ramp}). This reproduces, on our testbed, the known sequential-FDI evasion of forecasting-aided detection~\cite{zhao2017fase,manandhar2014kalman}. Ramp-rate and step-bias detectability are reported once, from the joint evaluation described next, in which the temporal, manifold and fused tests are calibrated together to a common false-alarm target, so that the three are directly comparable.

\begin{figure*}[t]\centering\includegraphics[width=\textwidth]{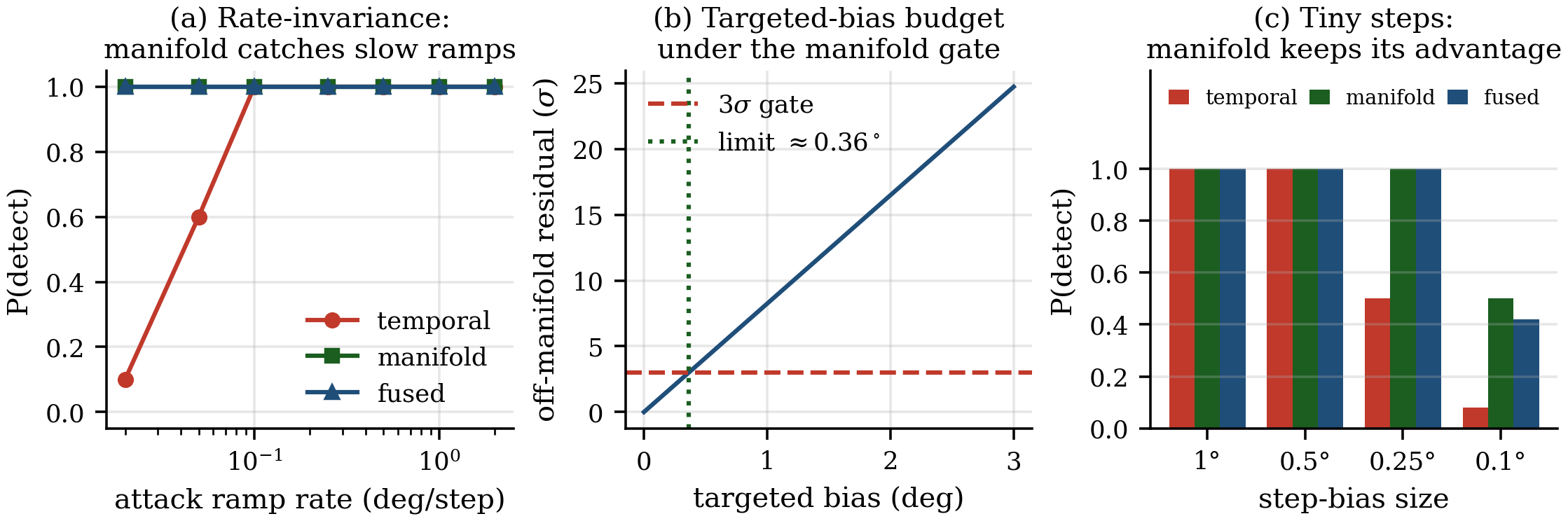}\caption{(a)~The manifold and fused detectors are rate-invariant, holding $\mathrm{P(detect)}{=}1.00$ (zero misses in $50$ trials, $\ge0.94$ at $95\%$ confidence) across every ramp rate tested, including $0.02^\circ$/step, where the temporal detector has collapsed to $0.10$. (b)~The off-manifold residual grows linearly with the injected bias; a $3\sigma_\rho$ spatial gate (three benign standard deviations of $\rho$) is crossed at $\approx0.36^\circ$. (c)~On tiny steps the manifold and fused detectors retain an advantage down to $0.10^\circ$ ($0.50$ and $0.42$ vs.\ $0.08$ temporal). The manifold is fitted on estimated states with $K{=}6$ chosen by benign cross-validation; measured held-out false-alarm rates are $4.5\%/5.3\%/5.0\%$ against a $5\%$ target.}\label{fig:exp6}\end{figure*}

A physics-manifold test closes the gap. Legitimate operating points occupy a low-dimensional, load-driven manifold whose dimension is selected by cross-validation on benign data alone, here $K{=}6$ of $27$ states. The manifold is fitted on the state estimates an operator actually observes, not on the true power-flow states: fitting on true states would supply ground-truth information unavailable in deployment, and on estimated states a $99.9\%$-cumulative-variance rule would instead return $K{=}20$ of $27$. The dimension is therefore set by a calibration criterion rather than by a variance threshold, and is not an arbitrary truncation of the $11$ PQ-bus state space. A targeted injection pushes the estimate off the manifold by an amount set by the \emph{bias}, not the ramp rate (Listing~\ref{lst:rho}, Eq.~\ref{eq:rho}). The off-manifold residual is therefore rate-invariant: it holds $\mathrm{P(detect)}{=}1.00$ across every ramp rate tested, exactly where the temporal test collapses (to $0.60$ at $0.05^\circ$/step and $0.10$ at $0.02^\circ$/step; Table~\ref{tab:ramp}, Fig.~\ref{fig:exp6}); on a slow $0.03^\circ$/step ramp the ROC separates cleanly (AUC temporal $0.72$ vs.\ manifold/fused $1.00$). All three tests run at their measured held-out false-alarm rates of $4.5\%$, $5.3\%$ and $5.0\%$ (temporal/manifold/fused), so the manifold advantage is not bought with extra false alarms. The manifold advantage extends to \emph{tiny} persistent steps as well: at a $0.25^\circ$ step the three detectors give $0.50/1.00/1.00$ and at $0.10^\circ$ they give $0.08/0.50/0.42$ for temporal/manifold/fused. At this smallest step the fused test falls slightly below the manifold test alone, so fusion does not always match the better single view. This manifold test also works within a single snapshot: its residual exceeds the $3\sigma_\rho$ gate once the targeted bias reaches about $0.36^\circ$, even though the residual-based bad-data tests remain blind to the same injection (Table~\ref{tab:acbdd}). This bounds the attacker's achievable targeted bias far below the temporal-only floor.

\begin{table}[t]\centering\caption{Detection probability vs.\ stealthy-FDI ramp rate at a common $5\%$ false-alarm rate (IEEE 14-Bus AC testbed). Slow ramps escape the temporal test; they do not escape the rate-invariant manifold test or fusion. Each cell is $50$ trials; an entry of $1.00$ means zero misses in $50$, i.e.\ $\mathrm{P(detect)}\ge0.94$ at $95\%$ confidence (rule of three). The benign manifold is fitted on \emph{estimated} states and its dimension $K{=}6$ is selected by cross-validation on benign data only. Thresholds are calibrated on $600$ benign streams; the false-alarm rates measured on $400$ held-out benign streams are $4.5\%/5.3\%/5.0\%$ (temporal/manifold/fused).}\label{tab:ramp}
\footnotesize\begin{tabular}{lccc}\toprule
Ramp (deg/step) & Temporal & Manifold & Fused\\\midrule
0.50 & 1.00 & 1.00 & 1.00\\
0.25 & 1.00 & 1.00 & 1.00\\
0.10 & 1.00 & 1.00 & 1.00\\
0.05 & 0.60 & \textbf{1.00} & \textbf{1.00}\\
0.02 & 0.10 & \textbf{1.00} & \textbf{1.00}\\\midrule
AUC @ $0.03$ & 0.72 & \textbf{1.00} & \textbf{1.00}\\\bottomrule\end{tabular}\end{table}

\begin{figure}[t]\centering\includegraphics[width=0.92\linewidth]{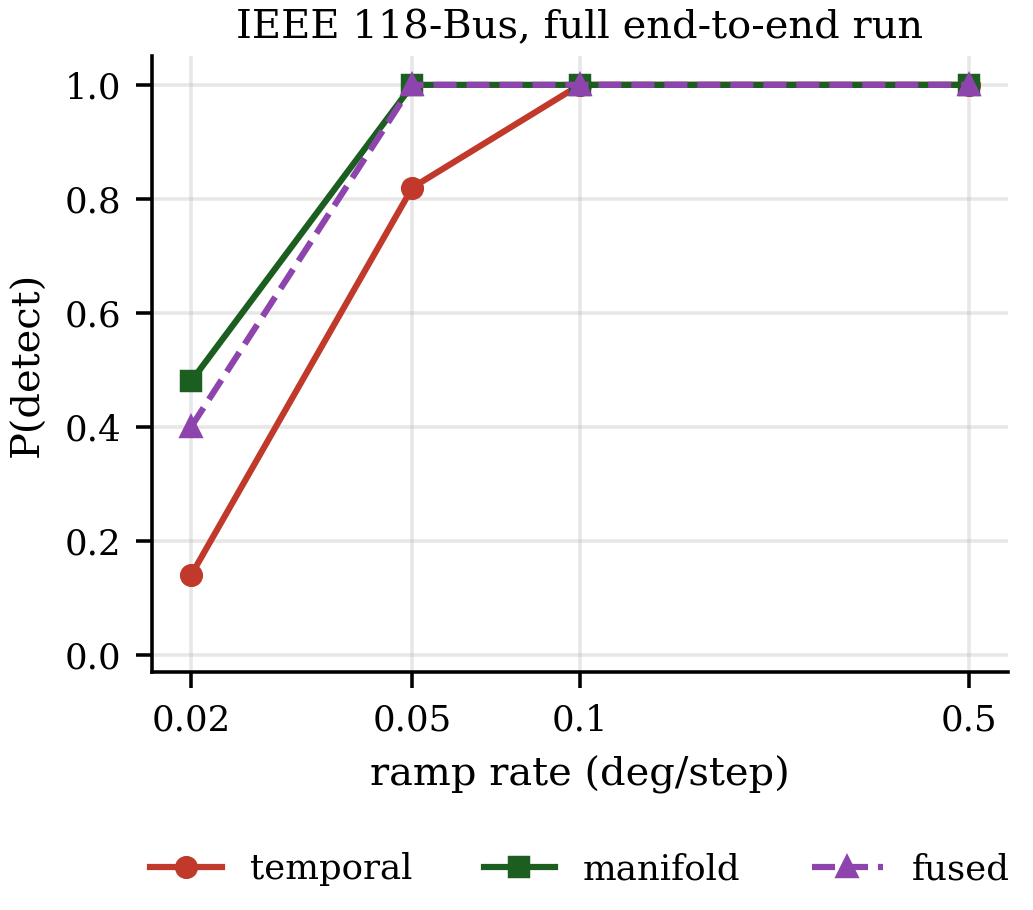}\caption{Full end-to-end IEEE 118-Bus validation with the estimator solved at every timestep. Detection probability against ramp rate, with every test calibrated to a common $5\%$ false-alarm target. The manifold statistic holds at unity where the temporal test begins to fail, and both degrade at the slowest rate tested.}\label{fig:ieee118full}\end{figure}

Under the full IEEE 118-Bus estimator the IEEE 14-Bus snapshot result above (Table~\ref{tab:acbdd}) reproduces exactly: an AC-consistent injection leaves the chi-square statistic unchanged ($417$ against a $539$ threshold) and the largest normalized residual unchanged ($2.95$, below the $3.0$ threshold), while the estimator settles at the intended $4.02^\circ$ bias. The detection comparison, with every test calibrated to a common $5\%$ false-alarm target, is given in Table~\ref{tab:ieee118full} and Fig.~\ref{fig:ieee118full}. For ramp rates of $0.05^\circ$/step and faster the manifold and fused detectors identify every attacked trial, while the temporal test has already begun to fail ($0.82$ at $0.05^\circ$/step). At the slowest rate tested, $0.02^\circ$/step, the temporal test collapses to $0.14$ and the manifold test falls to $0.48$. The manifold test reaches this higher detection rate while raising fewer false alarms than the temporal test ($1.1\%$ against $6.7\%$), so its advantage is not an artifact of calibration. The manifold test's false-alarm rate sits below the $5\%$ target because one alarm in the $90$ held-out benign streams is worth $1.1$ percentage points; this works against the manifold test, since a threshold relaxed to the full target could only raise its detection rate.

The manifold detection rate of $0.48$ at $0.02^\circ$/step is lower than the analytical prediction. An analytical calculation that assumes the consistent-injection identity holds exactly and that estimation noise is Gaussian predicts that the manifold test detects essentially every trial at $0.02^\circ$/step; the full simulation, in which neither assumption is imposed, detects roughly half. The qualitative conclusion is the same in both: the manifold statistic remains substantially more sensitive than the forecasting-aided temporal test in exactly the slow-ramp regime where the latter fails, by a factor of about $3.4$ at the slowest rate. The quantitative claim, however, is weaker on the larger system than on IEEE 14-Bus. Rate-invariance holds down to about $0.05^\circ$/step and then degrades, so on a system of this size an attacker who is willing to introduce the bias slowly enough retains a meaningful probability of evading both detectors. We regard this as the more useful result: it identifies the operating regime in which the approach is dependable, and the regime in which it is not.

\begin{table}[t]\centering\caption{Full end-to-end IEEE 118-Bus detection probability with every test calibrated to a common $5\%$ false-alarm target ($50$ trials per ramp rate). The achieved benign false-alarm rates are $0.067$/$0.011$/$0.056$ for temporal/manifold/fused, so the manifold test operates more conservatively than the other two. The lower bound is the one-sided $95\%$ binomial bound on the manifold detector.}\label{tab:ieee118full}
\footnotesize\begin{tabular}{@{}lcccc@{}}\toprule
Ramp (deg/step) & Temporal & Manifold & Fused & Manifold LB$_{95}$\\\midrule
$0.50$ & 1.00 & \textbf{1.00} & 1.00 & 0.94\\
$0.10$ & 1.00 & \textbf{1.00} & 1.00 & 0.94\\
$0.05$ & 0.82 & \textbf{1.00} & 1.00 & 0.94\\
$0.02$ & 0.14 & \textbf{0.48} & 0.40 & 0.34\\\bottomrule\end{tabular}\end{table}

The bias budget gives a consistent reading of the IEEE 14-Bus result. The off-manifold residual crosses its $3\sigma_\rho$ gate at about $0.36^\circ$ (Fig.~\ref{fig:exp6}b). At $0.02^\circ$/step the attacked 14-Bus stream accumulates about $0.46^\circ$ before it ends, which is above that gate, and this is consistent with the manifold test detecting every trial. No equivalent gate has been computed for IEEE 118-Bus, so the $0.48$ detection rate there is not yet explained by a measured budget. Computing that gate is the direct test of this explanation.

\begin{figure}[t]\centering\includegraphics[width=0.86\linewidth]{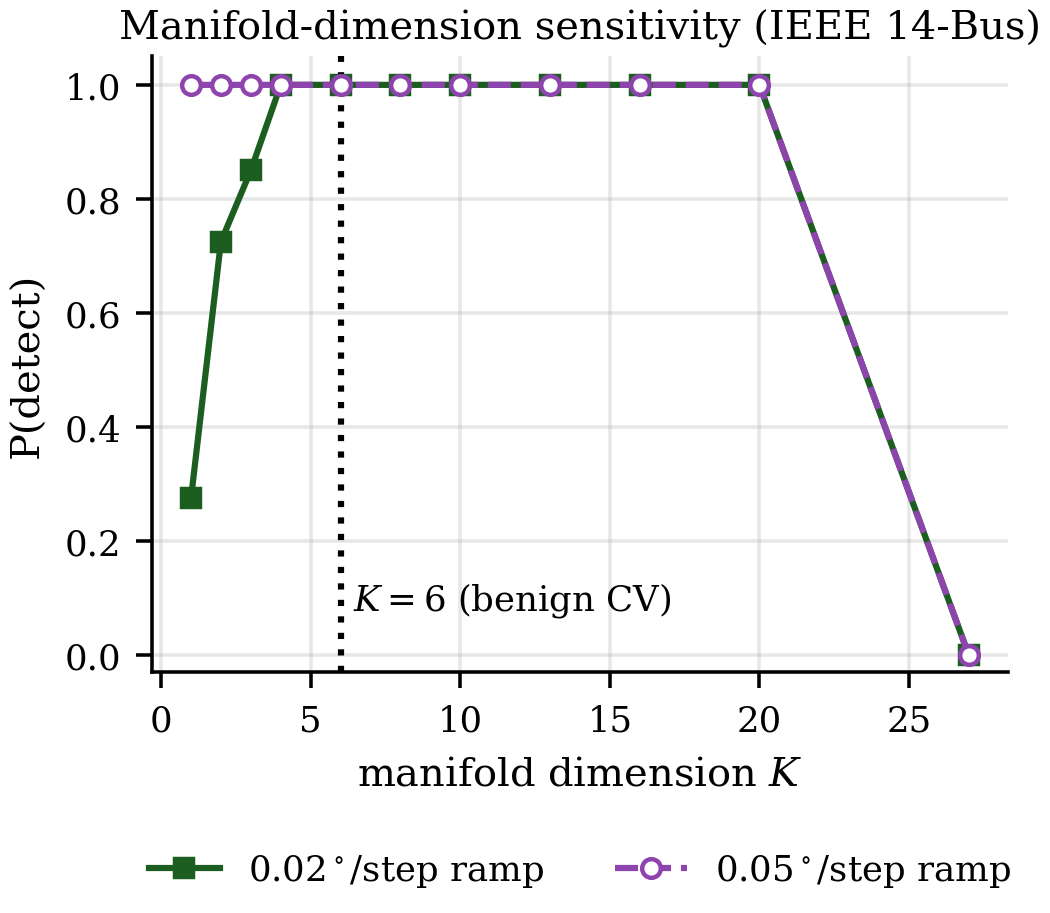}\caption{Manifold-dimension sensitivity on IEEE 14-Bus ($40$ trials/ramp, common $5\%$ false-alarm rate), with the manifold fitted on estimated states. Slow-ramp detection is flat for $K\in[4,20]$ and the cross-validated $K{=}6$ sits on the plateau; $K{\le}3$ underfits the benign load variation, and only at full rank $K{=}27$ do the surplus components absorb the attack direction and the residual go blind. Because each threshold is set on a finite benign set, the measured false-alarm rate varies between $0\%$ and $20\%$ across $K$; detection is still $1.00$ at $K{=}4$, $8$, $13$ and $20$, where the measured false-alarm rate is $0\%$, so the plateau is not produced by extra false alarms.}\label{fig:ksweep}\end{figure}

We can bound the manifold-aligned attacker directly. Because the benign manifold is only $K{=}6$-dimensional, a targeted single-coordinate bias has on average $88\%$ of its standardized energy (at least $63\%$ over coordinates, and $89\%$ for our target coordinate) forced off the manifold, so it cannot be concealed within the manifold-residual budget without abandoning the target coordinate (\texttt{exp9\_manifold\_attacker.py}). A sensitivity sweep over the manifold dimension confirms the choice is not delicate (Fig.~\ref{fig:ksweep}): at a common $5\%$ false-alarm rate, slow-ramp detection is flat for $K\in[4,20]$ (with the cross-validated $K{=}6$ on the plateau) and degrades only for $K{\le}3$, where the manifold underfits the benign load variation, and at full rank $K{=}27$, where the surplus components absorb the attack direction itself and the off-manifold residual can no longer see it (\texttt{exp6b\_ksweep.py}).

\section{Discussion}\label{sec:disc}
\subsection{Shuffling rewards the shortcut, not the model}
The TDNN is the most inflated ($0.99$) and the most penalized ($0.12$ drop). Shuffling puts adjacent windows on both sides of the split, so temporal capacity reads a near-copy instead of attack structure. The protocol gap therefore prices a shortcut all families used, not a weakness of the interpretable one. Across the five families shuffling inflates macro-F1 by $0.05$ to $0.12$. Under rolling-origin evaluation the four non-linear families differ from one another by only $0.03$, so the choice of protocol matters more than the choice among competitive models.

\subsection{Feasibility constrains the attacker, and the collapse survives anyway}
Feature-space attacks edit the detector inputs freely. The ES attacker is instead confined to the power-flow manifold, yet it still drives recall to $0.26$. The worst-case numbers are therefore not an artifact of an unconstrained threat model. Only feasible-space training recovers ($0.26$ to $0.53$), and only partly, because the feasible set is physics, not noise.

\subsection{The boundary is informational, so no better model crosses it}
Novelty is about zero for every attack, because the manipulated value lives in the payload and not in the headers. A quantity absent from the input cannot be recovered by any function of it. Commanded ground truth on IEEE 14-Bus reproduces the same partition, localizing the cause to the observable rather than the capture.

\subsection{Temporal and spatial tests fail on orthogonal axes}
The temporal test reads change over time, so a slow enough ramp is absorbed by the trend filter. The manifold residual reads distance from the benign subspace, set by accumulated bias, not by how it arrived. Evading both at once is the attacker's real constraint. A bias-budget effect is consistent with the IEEE 14-Bus results, but it has not been measured on IEEE 118-Bus, so it is offered as a hypothesis for the 118-Bus drop at $0.02^\circ$/step rather than as its established cause.

\subsection{Further observations}
Several themes connect the results. The selective policy turns fusion into an operational decision rule that, on both datasets, concentrates errors into an abstainable, low-confidence set (\ref{app:sel}). The cyber-only boundary is fundamental: semantic attacks are invisible in network headers even to a command-transition model, which is the core case for fusion. On robustness, the near-equivalence of PGD and random search on forests is gradient masking~\cite{papernot2016transferability,carlini2017robustness}; only the adaptive, joint, and ES attacks expose the true vulnerability, which is why we foreground them.

\subsection{Limitations}
The study also has clear limitations. Cyber and physics are never fused on a single realistic system: fusion is demonstrated on dataset~1, which exposes no topology and hence supports no real physics attacks; physics attacks are demonstrated on the simulated IEEE testbeds, whose cyber layer is simulated; and the IEC-104 study is cyber-only. The claim that fusion closes the observability boundary is therefore established piecewise across three disjoint testbeds and unified conceptually (Section~\ref{sec:resD3}) rather than end-to-end, and an aligned real-traffic, real-topology co-simulation is the natural next testbed.

The manifold detector itself uses a linear PCA fit and a synthetic load model, so nonlinear (kernel) manifolds, real load traces, a manifold-aligned adaptive attacker, self-supervised pretraining~\cite{selfsup2024}, and certified or state-space defenses are priorities. Finally, on novelty, the fusion model is standard; the contributions are the leakage-controlled interpretable baseline, the graded robustness study with its feasibility distinction, the observability taxonomy and cyber-only boundary, and the temporal-versus-spatial consistency study on a real-time AC estimator.

The manifold is a linear PCA fit on IEEE 14-Bus driven by a synthetic load model. Real daily and seasonal load traces would be expected to raise the manifold dimension somewhat, though it should remain low because operating conditions across buses are strongly correlated, so the off-manifold test should retain its behavior; confirming this on measured load data, together with larger systems, nonlinear manifolds, and a fully adaptive multi-bus attacker that spreads energy on-manifold, remains open.

\section{Conclusion}\label{sec:conc}
This study supports two claims.

\emph{Claim 1.} Shuffled, clean evaluation overstates security performance for every model family tested, by more than the competitive families differ from one another under honest evaluation, and physics-feasible evasion widens the gap further. On dataset~1, shuffling inflates macro-F1 by $0.05$ to $0.12$ across five model families, against a spread of only $0.03$ among the four non-linear families under rolling-origin evaluation, and on IEEE 14-Bus a physics-feasible evasion collapses FDI recall to $0.26$ under limited observability.

\emph{Claim 2.} Detectability is set by the observable, not the model: structural attacks need headers, naive value manipulation needs the residual, and stealthy FDI needs the state manifold, which on the larger IEEE 118-Bus system sees it reliably only above roughly $0.05^\circ$/step. Structural attacks are caught from network features and semantic value manipulation is not. A stealthy FDI is invisible to every snapshot view, but the rate-invariant manifold test detects it on IEEE 14-Bus at every ramp rate tested and on IEEE 118-Bus down to about $0.05^\circ$/step.

All code and both generated benchmarks are released under the INQUIRE Lab Academic Research License (code under the PolyForm Noncommercial License 1.0.0; generated data and figures under CC BY-NC 4.0) (Data availability). Future work: AC-feasible attacks at scale; certified and moving-target defenses~\cite{cohen2019certified,mtd2025}; topology-aware GNNs~\cite{sweeten2025fusion,boyaci2022gnn,kipf2017gcn}; and cross-testbed validation.

\section*{Reproducibility}
Code (fusion, selective detection, robustness, the five-family temporal baseline comparison \texttt{exp10\_temporal\_baselines.py} with an optional recurrent LSTM \texttt{exp10b\_lstm\_optional.py}, IEEE 14-Bus \texttt{pandapower} DC and full AC state estimation with temporal and manifold detectors \texttt{exp5\_ac\_stealth\_temporal.py}, \texttt{exp6\_manifold\_detector.py}, and IEC-104/MMS cross-validation) and all figures are released with this article. The rolling-origin temporal evaluation is regenerated by \texttt{exp8\_rolling\_origin.py}, and the manifold-attacker bound by \texttt{exp9\_manifold\_attacker.py}. The detector comparison (Table~\ref{tab:ramp}, Fig.~\ref{fig:exp6}) is produced at $50$ trials/point by \texttt{exp6\_submission.py} at a common $5\%$ false-alarm target, and the full IEEE 118-Bus validation (Table~\ref{tab:ieee118full}, Fig.~\ref{fig:ieee118full}) by \texttt{exp7c\_ieee118\_full.py}.

\section*{CRediT authorship contribution statement}
\textbf{Muhammad Ridwanul Hoque:} Conceptualization, Methodology, Software, Formal analysis, Investigation, Data curation, Writing -- original draft, Visualization. \textbf{Sarah Sharif:} Conceptualization, Methodology, Writing -- review \& editing, Supervision. \textbf{Reza Saeed Kandezy:} Validation, Writing -- review \& editing. \textbf{Yaser Banad:} Conceptualization, Writing -- review \& editing, Supervision, Project administration.

\section*{Declaration of competing interest}
The authors declare that they have no known competing financial interests or personal relationships that could have appeared to influence the work reported in this paper.

\section*{Funding}
This research did not receive any specific grant from funding agencies in the public, commercial, or not-for-profit sectors.

\section*{Declaration of generative AI and AI-assisted technologies in the manuscript preparation process}
During the preparation of this work the authors used Claude (Anthropic) in order to improve readability and language, to help reorganize the manuscript, and to assist with code for figure generation. After using this tool, the authors reviewed and edited the content as needed and take full responsibility for the content of the published article.

\section*{Data availability}
The datasets used in this study are publicly available (SmartGrid-CyberPhysical-Attack-Dataset on IEEE DataPort~\cite{sweeten2025dataset}; the IEC-104/IEC-61850 ICS traces of Matou\v{s}ek et al.~\cite{matousek2022ics}). All analysis code, generated benchmarks, and figure-reproduction scripts are released with this article under the INQUIRE Lab Academic Research License (code under the PolyForm Noncommercial License 1.0.0; generated data and figures under CC BY-NC 4.0) at \url{https://github.com/INQUIRELAB/smartgrid-cyberphysical-fusion}.

\appendix
\setcounter{figure}{0}
\setcounter{table}{0}
\section{Supporting results not tied to a claim}\label{app:sup}
This appendix reports supporting results that neither claim depends on: a topology-free graph control, the dataset-1 per-class results, and selective detection with error escalation.

\begin{figure}[!htb]\centering\includegraphics[width=\linewidth]{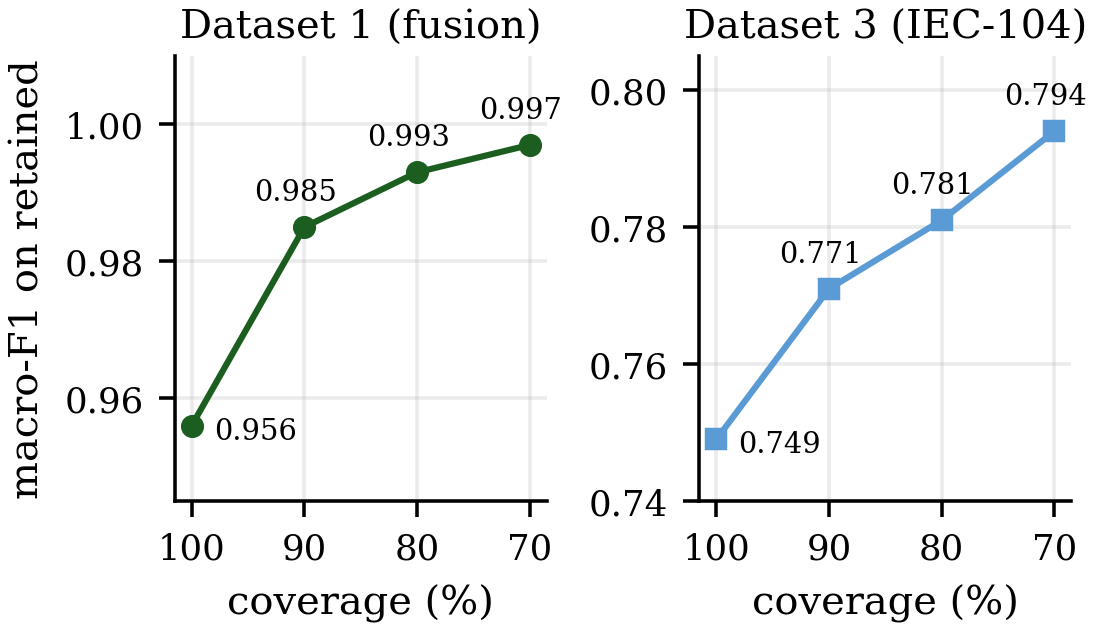}\caption{Selective detection on both datasets: abstaining on low-confidence windows (escalated to physical review) raises accuracy on the retained set.}\label{fig:sel}\end{figure}

\subsection{Topology-free graph control and per-class results}\label{app:gcn}
\begin{figure}[!htb]\centering\includegraphics[width=\linewidth]{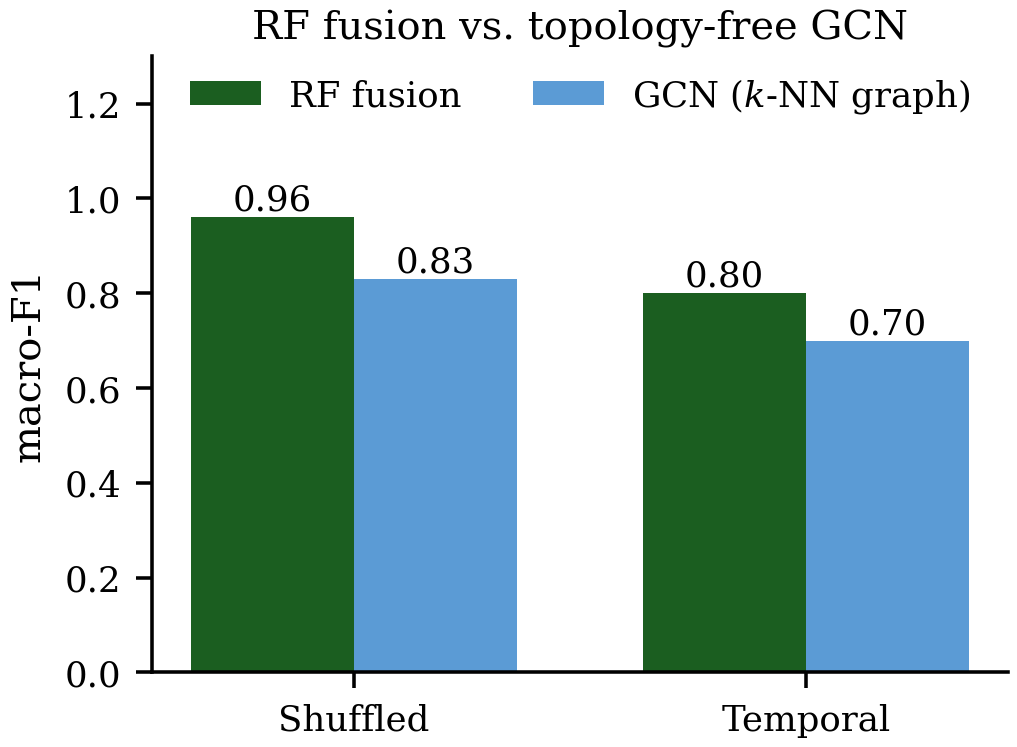}\caption{RF fusion vs.\ the topology-free GCN control over a $k$-NN feature graph ($k{=}10$), on identical splits. This is a sanity-check control built for this study, not the topology-aware fusion GNN of Sweeten et al.~\cite{sweeten2025fusion}.}\label{fig:g2g}\end{figure}

\begin{figure}[!htb]\centering\includegraphics[width=0.92\linewidth]{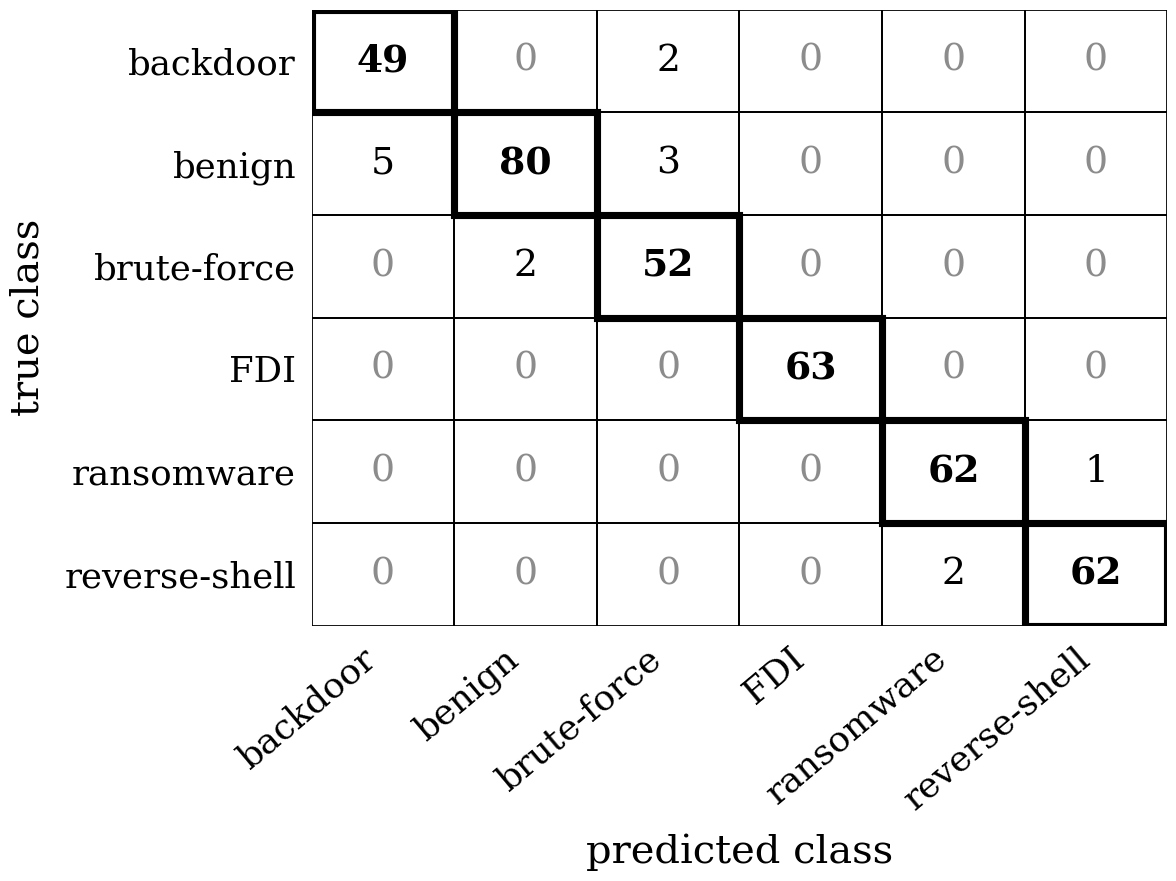}\caption{Dataset-1 confusion matrix for the fusion detector on the $75/25$ stratified split ($n{=}383$ test windows; rows are the true class, columns the predicted class; boxed bold cells on the diagonal are correct predictions). The residual errors are concentrated in two adjacent pairs: five benign windows are labelled backdoor, and a small number of windows are exchanged between ransomware and reverse-shell. Every FDI window is classified correctly, which is consistent with the strong marginal signature of this dataset's non-stealthy FDI generation discussed above.}\label{fig:cm}\end{figure}

A graph-model control serves as a sanity check, not a state-of-the-art comparison. Dataset~1 exposes no grid topology, so no topology-aware model can be run on it; the standard topology-free substitute, a symmetric $k$-nearest-neighbor ($k$-NN) graph ($k{=}10$) over standardized fused features, supplies the graph for a 2-layer graph convolutional network (GCN) trained for transductive node classification on the \emph{same} splits. The control asks one question: does graph structure add information when no real topology exists? It does not; the $k$-NN graph is built from the very features the RF already reads, so the RF's advantage ($0.96$ vs.\ $0.83$ shuffled; $0.80$ vs.\ $0.70$ temporal; Fig.~\ref{fig:g2g}) confirms that a lightweight, explainable model loses nothing in this topology-free setting. The fusion GNN of Sweeten et al.~\cite{sweeten2025fusion} is genuinely topology-aware (physical edges weighted by line admittance, cyber edges by router links, and explicit cyber-physical coupling edges) and reports substantially higher fusion F1 under its own protocol; it is not reimplemented here, the two numbers are not comparable, and no claim over it is made. A faithful topology-aware reimplementation under the temporal protocol remains future work. The Random Forest also matches gradient boosting and beats an MLP (Table~\ref{tab:det}); SHAP confirms reliance on both domains (Fig.~\ref{fig:shap}). The confusion matrix in Fig.~\ref{fig:cm} shows where the remaining errors fall. One per-class number deserves comment: FDI scores a perfect $1.00$ (Table~\ref{tab:perclass}) on only ${\sim}63$ test windows, so the $95\%$ upper bound on its true error rate is ${\approx}5\%$ (rule of three). Inspection shows why: this dataset's injections inflate the per-window current variance to ${\sim}11$ benign standard deviations, a strong marginal signature (though not a trivially separable one; a depth-2 tree reaches only $0.77$ F1 on FDI-vs-rest). The perfect score is thus a property of this dataset's non-stealthy FDI generation, not of FDI detection in general (the stealthy FDI of Sections~\ref{sec:resD2} and~\ref{sec:resD4} is the hard case), and it modestly flatters the fusion macro-average.

\begin{table}[!htb]\centering\caption{Dataset-1 per-class performance of the fusion detector on the $75/25$ stratified split ($383$ test windows; accuracy and macro-F1 $0.961$), the same split as Fig.~\ref{fig:cm}.}\label{tab:perclass}
\footnotesize\begin{tabular}{lccc}\toprule Class & Precision & Recall & F1\\\midrule
benign & 0.98 & 0.91 & 0.94\\ backdoor & 0.91 & 0.96 & 0.93\\
brute-force & 0.91 & 0.96 & 0.94\\ FDI & 1.00 & 1.00 & 1.00\\
ransomware & 0.97 & 0.98 & 0.98\\ reverse-shell & 0.98 & 0.97 & 0.98\\\bottomrule\end{tabular}\end{table}

\subsection{Selective detection and error escalation}\label{app:sel}
Errors are far less confident than correct predictions (mean $\kappa=0.65$ vs.\ $\kappa=0.93$), so abstaining on the least-confident $20\%$ of windows captures $88\%$ of all errors while retaining $80\%$ coverage at $0.993$ macro-F1 (Table~\ref{tab:sel1}); accuracy and macro-F1 coincide to three decimals in the table because the six classes are near-balanced. Methodologically this is standard selective prediction with a reject option; we claim the \emph{policy framing}, not a new mechanism. Note also that on dataset~1 the physical features are already inside the fused model, so ``escalation'' here means routing the abstained window to a \emph{human} analyst (or a slower verification process), not to a new observation layer; escalation to a genuinely new layer arises only in the cyber-only setting of Section~\ref{sec:resD3}.

\begin{table}[!htb]\centering\caption{Dataset-1 risk--coverage under selective abstention, computed from $5$-fold out-of-fold predictions over all $1{,}529$ windows. The $100\%$-coverage value ($0.956$) is therefore the $5$-fold figure and differs slightly from the $0.961$ of the single $75/25$ split in Table~\ref{tab:perclass}.}\label{tab:sel1}
\footnotesize\begin{tabular}{lcccc}\toprule Coverage & 100\% & 90\% & 80\% & 70\%\\\midrule
Accuracy & 0.956 & 0.985 & 0.993 & 0.997\\
macro-F1 & 0.956 & 0.985 & 0.993 & 0.997\\\bottomrule\end{tabular}\end{table}

Selective detection also helps here (Fig.~\ref{fig:sel}; macro-F1 $0.75\!\to\!0.79$). Crucially, the \emph{missed} semantic attacks are predicted benign with \emph{low} confidence ($91$--$96\%$ of the missed windows have confidence below $0.7$): the detector recognizes its own uncertainty on exactly the attacks it cannot classify; these are precisely the windows to escalate to the physical layer.

\bibliographystyle{elsarticle-num}
\bibliography{references}
\end{document}